\pdfoutput=1
\documentclass[11pt]{article}

\usepackage[preprint]{acl}

\usepackage{times}
\usepackage{latexsym}
\usepackage{amsmath}
\usepackage{amssymb}
\usepackage[T1]{fontenc}
\usepackage[utf8]{inputenc}

\usepackage{microtype}

\usepackage{inconsolata}

\usepackage{algorithm}
\usepackage{algorithmic}
\newcommand{\MYCOMMENT}[1]{/* #1 */}
\usepackage{bm}
\usepackage{graphicx}
\usepackage{multirow}
\usepackage{adjustbox}
\usepackage{booktabs}
\usepackage{colortbl}
\usepackage{tabularx}
\usepackage{tcolorbox}
\definecolor{pback}{RGB}{245,247,250}
\definecolor{pframe}{RGB}{30,60,120}
\definecolor{ptitle}{RGB}{50,50,50}
\title{Patterns Over Principles: The Fragility of Inductive Reasoning in LLMs under Noisy Observations}

\author{Chunyang Li, Weiqi Wang, Tianshi Zheng, Yangqiu Song \\
Department of Computer Science and Engineering, HKUST, Hong Kong SAR, China\\
\texttt{\{cliei, wwangbw, tzhengad, yqsong\}@cse.ust.hk}
}

\begin{document}
\maketitle
\begin{abstract}
Inductive reasoning, a cornerstone of human cognition, enables generalization from limited data but hasn't yet been fully achieved by large language models (LLMs). 
While modern LLMs excel at reasoning tasks, their ability to maintain stable and consistent rule abstraction under imperfect observations remains underexplored. 
To fill this gap, in this work, we introduce \textbf{Robust Rule Induction}, a task that evaluates LLMs' capability in inferring rules from data that are fused with noisy examples. To address this task,  we further propose Sample-steered Rule Refinement (SRR), a method enhancing reasoning stability via observation diversification and execution-guided feedback. 
Experiments across arithmetic, cryptography, and list functions reveal: (1) SRR outperforms other methods with minimal performance degradation under noise; 
(2) Despite slight accuracy variation, LLMs exhibit instability under noise (e.g., $0\%$ accuracy change with only $70\%$ consistent score);
(3) Counterfactual task gaps highlight LLMs' reliance on memorized patterns over genuine abstraction. 
Our findings challenge LLMs' reasoning robustness, revealing susceptibility to hypothesis drift and pattern overfitting, while providing empirical evidence critical for developing human-like inductive systems. Code and data are available at \href{https://github.com/HKUST-KnowComp/Robust-Rule-Induction}{https://github.com/HKUST-KnowComp/Robust-Rule-Induction}.
\end{abstract}

\section{Introduction}
Inductive reasoning\textemdash the cognitive capacity to generalize from specific instances to universal principles, is fundamental to human intelligence~\cite{Lake_Ullman_Tenenbaum_Gershman_2017}. 
This ability enables humans to abstract latent patterns from sparse data while maintaining robustness against conflicting evidence~\cite{Feldman1997TheSO}, as exemplified by children mastering linguistic rules despite exposure to occasional grammatical errors.
The robustness of inductive reasoning manifests in resistance to pattern interference (preserving learned rules when encountering contradictory examples) and tolerance to observational imperfections (maintaining stable performance under noisy learning conditions). Understanding and quantifying this robustness becomes increasingly crucial as artificial systems approach complex real-world applications where clean data remains elusive. 

Recent advances in large language models (LLMs) have demonstrated remarkable performance across various reasoning tasks~\cite{gpt4o,deepseekai2024deepseekv3technicalreport}, reigniting interest in comparing machine and human reasoning paradigms~\cite{collins_building_2024}. While contemporary studies showcase LLMs' proficiency in inductive reasoning, their capacity for robustness remains questionable. Unlike humans who can rapidly converge on correct rules through Bayesian hypothesis updating~\cite{doi:10.1126/science.1192788,doi:10.1126/science.aab3050}, current models often exhibit unstable reasoning trajectories when noise disrupts their thinking process~\cite{zhou2024can}. A growing body of evidence suggests that LLMs primarily engage in pattern matching rather than genuine reasoning~\cite{mirzadeh2024gsmsymbolicunderstandinglimitationsmathematical,wu-etal-2024-reasoning}.

Despite increasing attention to LLM inductive reasoning capabilities, current works exhibit limitations that obscure true inductive reasoning robustness. 
First, in prevailing and concurrent evaluations~\cite{pmlr-v139-alet21a,ijcai2024p693,zheng2025logidynamicsunravelingdynamicslogical}, models just predict outputs for novel inputs given exemplars, bypassing explicit rule verification. It fails to diagnose where and why rule induction fails. Existing studies that do generate intermediate rules either disregard noisy learning conditions~\cite{wang2024hypothesis,liu2024an} or assess performance solely through aggregate metrics like task accuracy gap~\cite{qiu2024phenomenal}. This neglects instance-level reasoning consistency\textemdash whether models maintain stable rule interpretations when exposed to conflicting patterns, which is a key indicator of human-like robust induction. Besides, prior methods often repurpose existing benchmarks (e.g., SCAN;~\citealt{pmlr-v80-lake18a}, ARC;~\citealt{chollet2019measureintelligence}) without modification, where potential data contamination may undermine validity as models could use memorized solutions.

To address these gaps, we introduce a novel task of robust rule induction, which challenges models to identify underlying functions from input-output exemplars containing controlled noise injections. Our approach features three innovations: (1) an evaluation pipeline from data synthesis to automatic evaluation, requiring models to output rules explicitly and enabling direct validation through programmatic execution. (2) A multi-dimensional robustness assessment with conventional accuracy and instance-level metrics. (3) A sample-steered refinement (SRR) methodology that improves both the capability and stability of LLMs' inductive reasoning through diverse sampling and iterative diagnostic refinement based on rule execution.
 
We curate three datasets spanning arithmetic, cryptography, and list operations\textemdash domains requiring progressively abstract rule formalization. The proposed SRR substantially improves performance across most tasks and maintains stability compared to other methods. However, comprehensive experiments reveal a critical dichotomy: while the accuracy variance achieves less than $5\%$, the consistency metric exposes fundamental instability. Furthermore, LLMs exhibit significant performance inconsistency with the same task category, performing considerably better on more familiar tasks than others, despite comparable complexity. Correct predictions may stem from recited patterns rather than stable rule abstraction. These findings challenge the ``real'' reasoning capabilities of LLMs.

\section{Preliminary}
\subsection{Problem Definitions}
\begin{figure}[ht]
    \centering
    \includegraphics[width=0.48\textwidth]{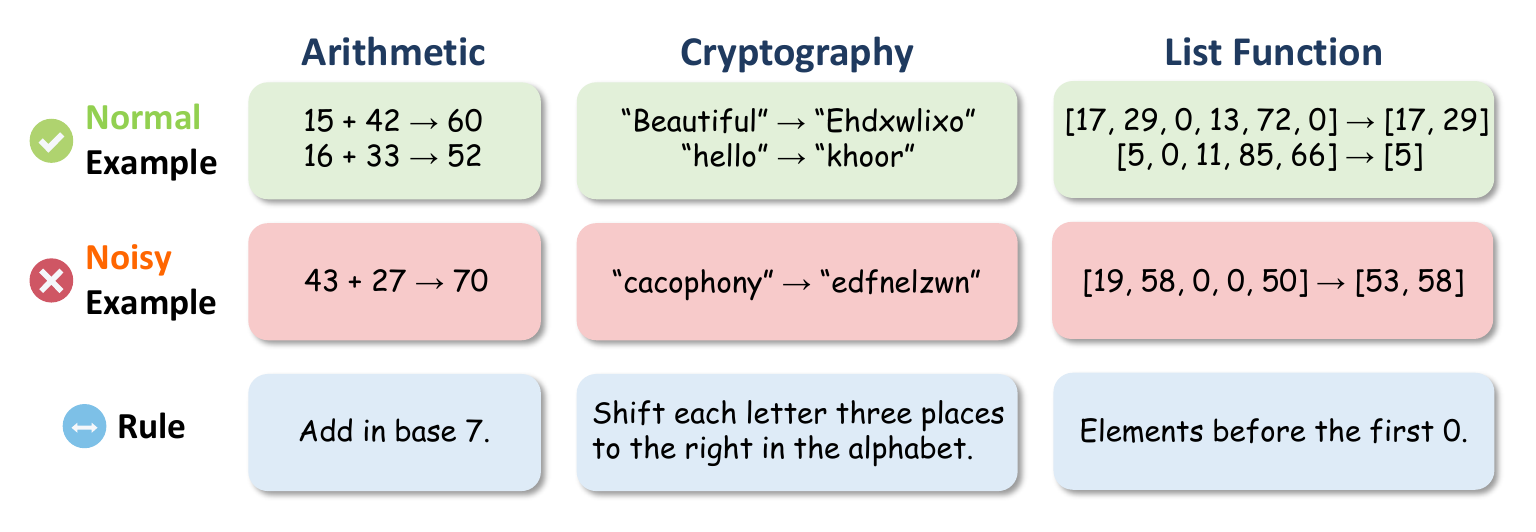}
    \caption{Example instances with noise and rules from Arithmetic$_{\text{base-7}}$, Cryptography$_{\text{Caesar}}$ and List Functions. }
    \label{fig:dataset_example}
\end{figure}

We formulate robust rule induction as the problem of identifying latent mapping functions from input-output pairs while maintaining consistent reasoning performance under noisy conditions. 
While prior works~\cite{wang2024hypothesis,liu2024an} focus on basic rule discovery capabilities, we specifically investigate the robustness of inductive reasoning when the observations contain conflicting patterns.
Some examples are shown in Figure~\ref{fig:dataset_example}. The observations may contain both normal and noisy examples, and the model is required to infer the correct rule regardless of the noise.

Formally, given a dataset $\mathcal{D}=\{(x_i,y_i)\}_{i=1}^N$ where $x\in\mathcal{X}$ denotes inputs and $y\in\mathcal{Y}$ represents outputs generated by an underlying function $f:\mathcal{X}\rightarrow\mathcal{Y}$, the objective is to induce an approximation $\hat{f}$ from observed examples $\mathcal{D}_{\text{seen}}\subseteq\mathcal{D}$. 
Robust rule induction should satisfy: $\hat{f}(x)=f(x)$ for $(x,y)\in \mathcal{D}_{\text{unseen}}$ despite $\mathcal{D}_{\text{seen}}$ containing noisy samples $\mathcal{D}_{\text{noise}}$ where $y\neq f(x)$.

\begin{figure*}[t]
    \centering
    \includegraphics[width=0.95\textwidth]{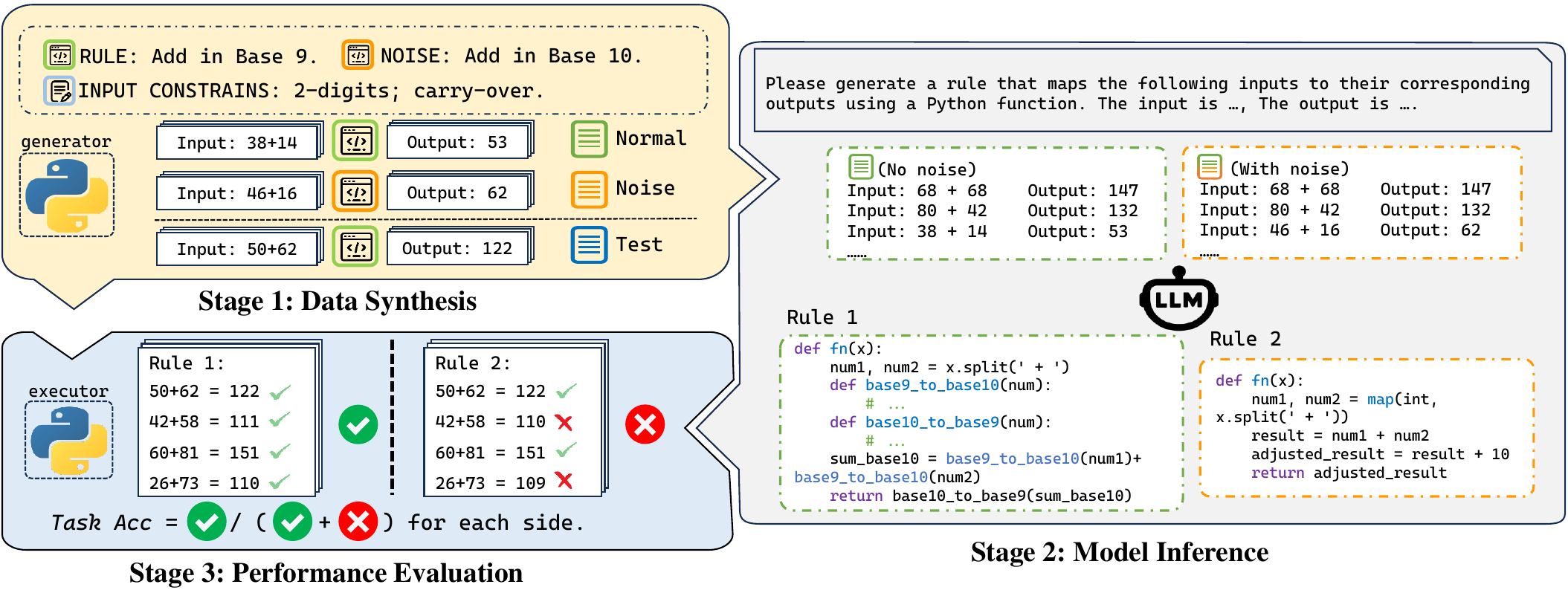}
    \caption{Evaluation pipeline exemplified by base-9 addition, consisting of three stages: (1) Data Synthesis, generating normal, noisy and test examples; (2) Model Inference, prompting models with seen examples to induce rules in Python function form; (3) Performance Evaluation, executing induced rules on test examples to assess correctness and robustness under noise.}
    \label{fig:evaluation_pipeline}
\end{figure*}
\subsection{Evaluation Pipeline}
As illustrated in figure~\ref{fig:evaluation_pipeline}, to evaluate the robustness of inductive reasoning in large language models, we propose a novel evaluation pipeline consisting of three main stages: data synthesis, model inference, and performance evaluation. In the data synthesis stage, due to the characteristics of our data, normal examples, noisy examples, and test examples for each instance in the dataset can be automatically generated by programs given the rule and noise definitions. The first two are mixed to form seen examples, while the latter is used to evaluate the model's reasoning ability. In the model inference stage, we prompt the language model with seen examples in standard input-output (IO) format, and the induced rules are restricted to Python function form for automatic evaluation. Finally, we evaluate the rule by executing it on the test examples. The instance is considered solved if all the test examples are correctly solved. More details about the evaluation pipeline, like the data synthesis process, can be found in appendix~\ref{sec:app_eval}.

We use task accuracy, which is the proportion of solved instances over the total number of instances, and its change under different conditions as the evaluation metrics.  While task accuracy is a comprehensive metric to evaluate the model's reasoning ability, it may not fully capture the robustness of the model at the instance level. To address this, we introduce the consistency score, which is defined as follows:
\begin{equation*}
    \text{Consistency Score} := \frac{1}{N}\sum_{i=1}^N\mathbb{I}[Sol_i^{\text{c}}=Sol_i^{\text{n}}]
\end{equation*}
where $N$ is the total number of instances in the dataset, $\mathbb{I}[\cdot]$ is the indicator function, $Sol_i^{\text{c}}$ is whether the $i$-th instance is solved without noise, $1$ if solved and $0$ otherwise, and $Sol_i^{\text{n}}$ is whether the $i$-th instance is solved with noise. This metric quantifies the model's ability to maintain stable reasoning conclusions under noise and offers a more granular view of the model's robustness.

\section{Sample-steered Rule Refinement}
Motivated by previous methods~\cite{qiu2024phenomenal,wang2024hypothesis} using iterative hypothesis refinement, we propose Sample-steered Rule Refinement (SRR), a framework to improve inductive reasoning in both clean and noisy conditions. This method conducts inductive reasoning through three-phase optimization: (1) \textit{contrastive hypothesis generation} to bootstrap diverse rule candidates, (2) \textit{diagnostic feedback construction} through evidence-aware sampling, and (3) \textit{iterative rule refinement} using failure-driven corrections. The full algorithm is detailed in Algorithm~\ref{alg:srr}.

\begin{algorithm}[ht]
\small
\caption{SRR Framework}
\label{alg:srr}
\begin{algorithmic}
\REQUIRE Seen examples $\mathcal{D}_{\text{seen}}$, LLM $M$, max iterations $T$, max subsets $K$, threshold $\tau$
\ENSURE Best hypothesis $\hat{f}^*$ (in Python function form)
\STATE $\mathcal{H}_0 \leftarrow \emptyset$
\FOR{$k = 1$ \TO $K$}
    \STATE $\mathcal{D}_k \leftarrow \text{SampleSubset}(\mathcal{D}_{\text{seen}})$
    \STATE $\mathcal{H}_0 \leftarrow \mathcal{H}_0 \cup \{M(\mathcal{D}_k, \text{"Generate rule function"})\}$
\ENDFOR
\STATE $\mathcal{H}_0 \leftarrow \mathcal{H}_0 \cup \{M(\mathcal{D}_{\text{seen}}, \text{"Generate rule function"})\}$

\STATE $\hat{f}_0 \leftarrow \arg\max_{h\in\mathcal{H}_0} \text{Acc}(h, \mathcal{D}_{\text{seen}})$

\FOR{$t = 1$ \TO $T$}
    \IF{$\text{Acc}(\hat{f}_{t-1}, \mathcal{D}_{\text{seen}}) \geq \tau$}
        \RETURN $\hat{f}_{t-1}$
    \ENDIF
    \STATE $\mathcal{C}_t \leftarrow \text{CorrectExamples}(\hat{f}_{t-1}, \mathcal{D}_{\text{seen}})$
    \STATE $\mathcal{E}_t \leftarrow \text{WrongExamples}(\hat{f}_{t-1}, \mathcal{D}_{\text{seen}})$
    \STATE $\text{Feedback} \leftarrow \{\mathcal{C}_t[:n], \mathcal{E}_t[:m]\}$ \MYCOMMENT{Sample n,m cases}
    
    \STATE $h_{\text{new}} \leftarrow M(\hat{f}_{t-1}, \text{Feedback}, \text{"Revise rule"})$
    \IF{$\text{Acc}(h_{\text{new}}, \mathcal{D_{\text{seen}}}) > \text{Acc}(\hat{f}_{t-1}, \mathcal{D_{\text{seen}}})$}
        \STATE $\hat{f}_t \leftarrow h_{\text{new}}$
    \ELSE
        \STATE $\hat{f}_t \leftarrow \hat{f}_{t-1}$
    \ENDIF
\ENDFOR

\RETURN $\hat{f}^* = \arg\max_{t}\text{Acc}(\hat{f}_t, \mathcal{D}_{\text{seen}})$
\end{algorithmic}
\end{algorithm}

\paragraph{Diversity-aware Hypothesis Generation} Different from previous methods based on the whole seen examples, we first generate $K$ hypotheses from random subsets of the seen examples, plus one hypothesis using the full observations. This step ensures coverage of both local patterns and global consistency. The number of examples in the subset is less than the full seen set, thus the noise in the subset has a greater impact. The sampled subset may also contain completely noise-free examples, which can better distinguish the rules.

\paragraph{Diagnostic Feedback Construction} After generating the initial hypotheses, the induced rule is applied to the seen examples through code interpreter. We collect the correct and incorrect cases and sample these cases as feedback. We focus more on the incorrect cases, as they provide more information about the rule refinement, while the correct cases are used as positive feedback to reinforce the rule. The hypothesis with the highest accuracy on the seen examples is selected as the initial rule.

\paragraph{Iterative Rule Refinement} At each iteration, the model receives the current selected rule with sampled feedback to generate a new rule. The new hypothesis is compared to the previous one, and the more accurate one is selected. This iterative process continues until the accuracy of the rule on the seen examples exceeds a predefined threshold or the maximum number of iterations is reached.

\section{Experiments}
In this section, we evaluate the robustness of inductive reasoning in language models on different tasks. We also compare the performance of the models under different noise levels and analyze the effectiveness of the different methods in enhancing the robustness of inductive reasoning.

\subsection{Experimental Setup}
Consistent with previous studies~\cite{wang2024hypothesis,qiu2024phenomenal}, we adopt few-shot prompting to assess the models' inductive capabilities. Each instance contains $10$ normal examples, $5$ noisy examples, and $10$ test examples. The normal and noisy examples are mixed to form $10$ seen examples, which serve as prompts. This approach preserves task semantics while introducing controlled perturbations, thereby simulating real-world scenarios where observational data often contains inherent imperfections. LLMs are explicitly informed that examples may contain some noise. We formulate the output of LLMs as Python functions and execute them on the test examples to automatically evaluate the inferred rules. More details about the experiments can be found in the appendix~\ref{sec:app_exp}.

\subsection{Datasets}
We use three datasets with their subsets: Arithmetic, Cryptography, and List Functions. These datasets offer different rule induction challenges, including mathematical calculations, symbol representation, and list operations. To ensure that the noise reflects plausible perturbations encountered in real-world scenarios, we balance realism with experimental control while minimizing artificiality in designing the noise. Comprehensive description and statistics are shown in Table~\ref{tab:data_statistic}.

\begin{table}[ht]
    \small
    \centering
    \begin{adjustbox}{max width=1\linewidth}
    {
    \begin{tabular}{c|l|c|c}
    \toprule
    \textbf{Dataset} & \textbf{Noise Construction} & \textbf{Subset} & \# \textbf{Instance} \\
    \midrule
    \multirow{3}{*}{Arith.} & \multirow{3}{*}{Decimal Addition} & 7-base & $100$ \\
     &  & 8-base & $100$ \\
     &  & 9-base & $100$ \\
    \midrule
    \multirow{3}{*}{Crypto.} & \multirow{3}{*}{Random Replacement} & Caesar & $100$ \\
     &  & Atbash & $100$ \\
     &  & Keyboard & $100$ \\
    \midrule
    List Func. & Random Replacement & N/A & $250$ \\
    \bottomrule
    \end{tabular}
    }\end{adjustbox}
    \caption{Overview of datasets, including noise construction methods, subsets, and the number of instances per dataset. Arith., Crypto., List Func. denote Arithmetic, Cryptography and List Functions respectively.}
    \label{tab:data_statistic}
\end{table}

\paragraph{Arithmetic} 
The arithmetic task, introduced by~\citet{wu-etal-2024-reasoning}, involves counterfactual two-digit addition in non-decimal bases. To modulate difficulty and prioritize reasoning over memorization, we focus on the base-7, 8, 9 systems, and the noisy example is the common 10-based equations. The input is two two-digit numbers in the corresponding base, connected by a plus sign, and the output is the sum of the two numbers. To ensure the model can reason correctly, we guarantee that there must be a carry-over in the addition process so that the model must reason about the carry-over instead of simply adding the numbers. 

\paragraph{Cryptography} 
We use three types of substitution ciphers: Caesar, Atbash, and Keyboard. The Caesar cipher is a cipher that shifts the alphabet by a fixed number of positions. The Atbash cipher is a cipher that replaces each letter with the letter symmetrically opposite in the alphabet (e.g., A to Z). The Keyboard cipher replaces letters according to their positions in the alphabet with the corresponding positions on the keyboard (e.g., A to Q, B to W). The input is a word, and the output is the encrypted text. We randomly replace the characters in the output with other characters to generate noisy examples.

\paragraph{List Functions} 
The list functions dataset~\cite{rule2020child} evaluates the concept learning ability in the domain of cognitive science. The task is to induce a function that maps a list of numbers to another list of numbers. Each instance in the dataset corresponds to a list manipulation operation, such as sorting, reversing, or filtering. We generate the examples of each instance under controlled conditions to ensure the rule can be induced. The noisy examples are generated by randomly replacing the numbers in the output list with other numbers.

\begin{figure*}[t]
    \centering
    \includegraphics[width=0.95\textwidth]{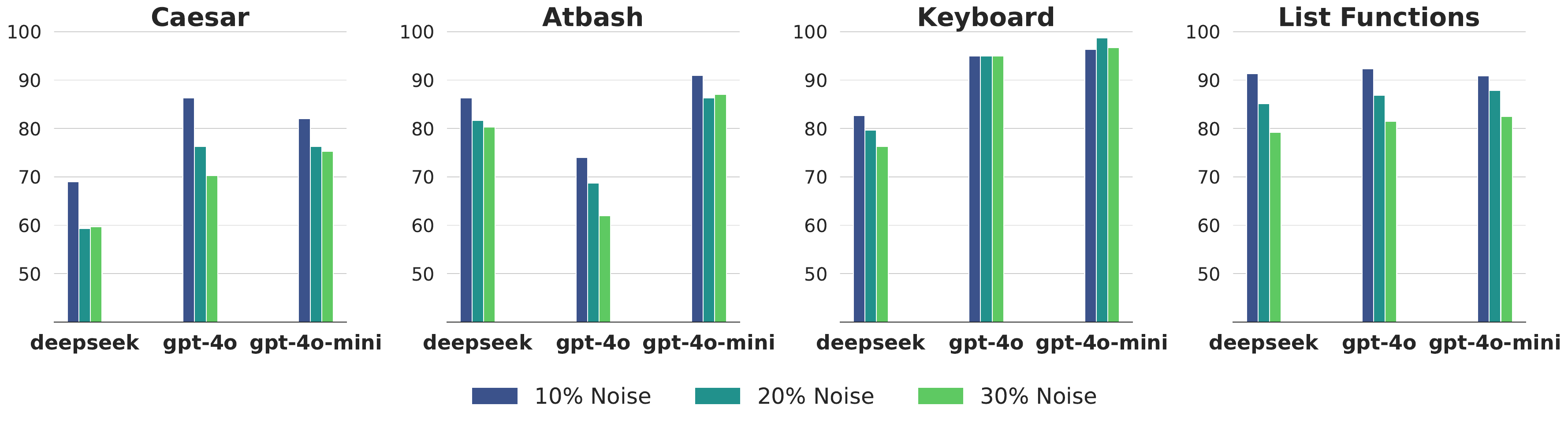}
    \caption{Consistency score$(\%)$ with clean data of different models on the Cryptography and List Functions datasets under different noise levels.}
    \label{fig:consistency_score}
\end{figure*}

\begin{table}[ht]
\small
\centering
\begin{adjustbox}{max width=1\linewidth}
{
\begin{tabular}{l|ccc|c}
\toprule
\multirow{2}{*}{\textbf{Model}} & \multicolumn{3}{c|}{\textbf{Cryptography}} & \multirow{2}{*}{\textbf{List Functions}} \\
\cmidrule(l){2-4}
&  Caesar & Atbash & Keyboard & \\
\midrule
GPT-4o-mini & $28.7_{_{\pm1.2}}$ & $6.3_{_{\pm1.2}}$ & $0.0_{\pm0.0}$ & $31.3_{\pm0.2}$ \\
 \hspace{10pt}\texttt{10\% noise} & $32.0_{\pm2.4}$&$12.7_{\pm2.5}$& $3.7_{\pm0.5}$ & $28.4_{\pm1.4}$ \\
 \hspace{10pt}\texttt{20\% noise} & $29.0_{\pm0.8}$&$13.3_{\pm2.6}$&$1.3_{\pm1.2}$& $23.5_{\pm1.0}$ \\
 \hspace{10pt}\texttt{30\% noise} & $21.3_{\pm0.9}$&$16.0_{\pm0.0}$&$3.3_{\pm0.5}$& $19.7_{\pm0.2}$ \\
\midrule
GPT-4o & $68.3_{\pm0.5}$ & $48.0_{\pm2.2}$ & $4.7_{\pm0.5}$ & $39.9_{\pm0.5}$ \\
 \hspace{10pt}\texttt{10\% noise} &$75.3_{\pm0.5}$&$34.7_{\pm0.9}$&$3.7_{\pm0.5}$& $36.9_{\pm1.0}$ \\
 \hspace{10pt}\texttt{20\% noise} &$75.3_{\pm2.1}$&$27.3_{\pm0.5}$&$1.7_{\pm0.9}$& $31.9_{\pm0.8}$ \\
 \hspace{10pt}\texttt{30\% noise} &$64.7_{\pm4.7}$&$22.7_{\pm2.4}$&$2.3_{\pm1.2}$& $26.1_{\pm0.5}$ \\
\midrule
DeepSeek-V3 &  $50.0_{\pm1.6}$ & $21.7_{\pm2.5}$ & $25.0_{\pm0.8}$ & $43.1_{\pm0.4}$ \\
 \hspace{10pt}\texttt{10\% noise} & $24.3_{\pm1.7}$ & $15.3_{\pm2.4}$ & $15.7_{\pm1.2}$ & $40.8_{\pm0.9}$ \\
 \hspace{10pt}\texttt{20\% noise} & $15.3_{\pm3.7}$ & $21.3 _{\pm2.5}$ & $11.3_{\pm1.2}$ & $35.3_{\pm 1.9}$ \\
 \hspace{10pt}\texttt{30\% noise} & $11.0_{\pm0.8}$ & $16.7_{\pm0.5}$ & $11.3_{\pm3.1}$ & $30.5_{\pm0.7}$ \\
\bottomrule
\end{tabular}
}
\end{adjustbox}
\caption{Task accuracy ($\%$) on Cryptography (including Caesar, Atbash, and Keyboard subtasks) and List Functions under different noise levels (proportion of noise in seen examples). Results are shown in mean $\pm$ standard deviation over 3 independent runs.}
\label{tab:robust_noise_level}
\end{table}

\subsection{Robustness Under Different Noise Levels}
\label{subsec:noise_level}
We first evaluate the robustness of inductive reasoning in language models under different noise injection ratios. Noise levels are defined as the proportion of noisy examples in the seen examples. We test three representative models: GPT-4o-mini~\cite{gpt4omini}, GPT-4o~\cite{gpt4o}, and DeepSeek-V3~\cite{deepseekai2024deepseekv3technicalreport} and ask the models to directly infer the rules from the seen examples without any additional output. Since the models cannot solve the Arithmetic task under the Direct Output (DO) setting, we present the results for the Cryptography and List Functions. The results are shown in Table~\ref{tab:robust_noise_level}. To better evaluate the robustness of the models, we further investigate the consistency score with clean data under different noise levels, as shown in Figure~\ref{fig:consistency_score}. According to the results, we have the following observations.

First, contrary to conventional assumptions, \textbf{noise introduction does not universally degrade performance, instead, the models exhibit performance fluctuations}, demonstrating their inherent sensitivity to conflicting patterns. For example, GPT-4o achieves improved accuracy on Caesar cipher tasks at $10\%$ noise ($7.0\%$ absolute improvement over clean data), while List Functions exhibit monotonic performance decay with increasing noise levels. In some cases, moderate noise even improves performance.

With the exception of tasks where models fundamentally struggle, experimental results demonstrate \textbf{a decline in consistency scores as noise levels escalate, and it declines more sharply than its task accuracy variation}. This discrepancy indicates noise introduces bidirectional reasoning instability: models not only fail on previously solvable instances (noise interference) but also succeed on originally challenging cases (incorrect generalization), leading to different performance changes.

\begin{table*}[ht]
\small
\centering
\begin{adjustbox}{max width=1\linewidth}
{
\begin{tabular}{l|l|ccc|ccc|c}
\toprule
\multirow{2}{*}{\textbf{Model}} & \multirow{2}{*}{\textbf{Method}} & \multicolumn{3}{c|}{\textbf{Arithmetic}} & \multicolumn{3}{c|}{\textbf{Cryptography}} & \multirow{2}{*}{\textbf{List Functions}} \\
\cmidrule(l){3-8}
 & & 7-base & 8-base & 9-base & Caesar & Atbash & Keyboard & \\
\midrule
\multirow{5}{*}{GPT-4o} & DO & $0.0$ & $0.0$ & $0.0$ & $75.3(\uparrow 7.0)$ & 
$34.7(\downarrow 13.3)$ & $3.7(\downarrow 1.0)$ & $36.9(\downarrow 3.0)$ \\
& CoT & $3.0 (\leftrightarrow \underline{0.0})$ & $8.0 (\downarrow 14.0)$ & $8.0(\uparrow 5.0)$ & $\textbf{85.0} (\uparrow 1.0)$ & $20.0(\downarrow 7.0)$ & $4.0 (\leftrightarrow \underline{0.0})$ & $40.4(\downarrow 5.2)$ \\
& SC & $0.0 (\downarrow 1.0)$ & $6.0 (\downarrow 7.0)$ & $0.0 (\downarrow 3.0)$ & $\textbf{85.0} (\leftrightarrow \underline{0.0})$ & $29.0 (\downarrow 11.0)$ & $5.0 (\downarrow 2.0)$ & $42.4 (\downarrow 3.2)$ \\
& SR & $1.0 (\downarrow 4.0)$ & $18.0 (\downarrow \underline{4.0})$ &$4.0 (\downarrow 2.0)$ & $81.0 (\downarrow 2.0)$ & $23.0 (\downarrow 8.0)$ & $5.0 (\leftrightarrow \underline{0.0})$ & $39.6 (\downarrow 4.4)$\\
& HR & $1.0 (\downarrow 4.0)$ & $28.0 (\leftrightarrow \underline{0.0})$ &$\textbf{19.0} (\downarrow \underline{1.0})$ & $84.0 (\leftrightarrow 0.0)$ & $\textbf{56.0} (\uparrow \underline{1.0})$ & $\textbf{22.0} (\uparrow 2.0)$ & $55.6 (\downarrow 2.0)$\\
\cmidrule{2-9}
& SRR & $\textbf{5.0} (\downarrow 1.0)$ & $\textbf{51.0}(\uparrow 6.0)$ & $\textbf{19.0}(\downarrow 2.0)$ & $\textbf{85.0} (\uparrow 3.0)$ & $52.0(\downarrow 3.0)$ & $9.0 (\uparrow 1.0)$ & $\textbf{57.2} (\downarrow \underline{1.6})$\\
\midrule
\multirow{6}{*}{Deepseek-V3} & DO & $0.0$ & $0.0$ & $0.0$ & $24.3(\downarrow 25.7)$ & $15.3(\downarrow 6.4)$ & $15.7(\downarrow 9.3)$ & $40.8(\downarrow \underline{2.3})$ \\
& CoT & $77.5(\downarrow 6.0)$ & $83.0(\downarrow 13.0)$ & $67.5(\downarrow 14.0)$ & $80.5(\downarrow 4.0)$ & $26.0(\downarrow 5.5)$ & $5.0(\downarrow 8.5)$ & $52.0(\downarrow 6.4)$ \\
& SC & $83.0(\downarrow 3.0)$ & $93.0 (\downarrow 6.0)$ & $81.0(\downarrow 3.0)$ & $\textbf{86.0}(\downarrow \underline{1.0})$ & $40.0 (\downarrow 3.0)$ & $7.0(\downarrow 4.0)$ & $56.0(\downarrow 3.2)$ \\
& SR &$70.0 (\downarrow 10.0)$ &$74.0 (\downarrow 9.0)$ &$68.0 (\uparrow 6.0)$ & $72.0(\downarrow 10.0)$ & $19.0 (\downarrow 1.0)$ & $8.0 (\downarrow 3.0)$ & $47.2 (\downarrow 10.0)$\\
& HR &$82.0 (\downarrow 4.0)$ &$89.0 (\downarrow \underline{2.0})$ &$78.0 (\leftrightarrow \underline{0.0})$ & $\textbf{86.0}(\leftrightarrow \underline{0.0})$ & $\textbf{56.0} (\leftrightarrow \underline{0.0})$ & $\textbf{17.0} (\downarrow 7.0)$ & $63.6 (\downarrow 6.0)$\\
\cmidrule{2-9}
& SRR & $\textbf{96.0}(\downarrow \underline{1.0})$ & $\textbf{95.0}(\downarrow 4.0)$ & $\textbf{94.0}(\leftrightarrow \underline{0.0})$ & $\textbf{86.0}(\downarrow 1.0)$ & $52.0(\downarrow 1.0)$ & $11.0(\downarrow \underline{2.0})$& $\textbf{64.8}(\downarrow 2.8)$\\
\bottomrule
\end{tabular}
}
\end{adjustbox}
\caption{Task accuracy ($\%$) on different datasets under $10\%$ noise. The numbers in parentheses are the change compared to the clean data, and the arrows indicate the direction of the change. \textbf{Bold} indicates the best performance, and \underline{underline} indicates the smallest change.}
\label{tab:methods_res}
\end{table*}

\begin{table}[ht]
\small
\centering
\begin{adjustbox}{max width=1\linewidth}
{
\begin{tabular}{l|rrr|rrr}
\toprule
\multirow{2}{*}{\textbf{Dataset}} & \multicolumn{3}{c|}{\textbf{GPT-4o}} & \multicolumn{3}{c}{\textbf{DeepSeek-V3}} \\
\cmidrule(l){2-7}
& SC & HR-0 & SRR-0 & SC & HR-0 & SRR-0 \\
\midrule
Arithmetic$_\text{7}$ & $0.0$ & $0.0$ & $4.0$ & $83.0$ & $81.0$ & $95.0$ \\
Arithmetic$_\text{8}$ &$6.0$ & $11.0$ & $40.0$ &$93.0$ & $85.0$ & $95.0$ \\
Arithmetic$_\text{9}$ &$0.0$ & $4.0$ & $15.0$ &$81.0$ & $68.0$ & $91.0$\\
Crypto$_{\text{Caesar}}$ &$85.0$ & $84.0$ & $85.0$ &$86.0$ & $86.0$ & $86.0$\\
Crypto$_{\text{Atbash}}$ &$29.0$ & $42.0$ & $51.0$ &$40.0$ & $52.0$ & $48.0$ \\
Crypto$_{\text{Keyboard}}$ &$5.0$ & $14.0$ & $8.0$ &$7.0$ & $10.0$ & $9.0$ \\
List Functions &$42.4$ & $48.8$ & $54.0$ & $56.0$ & $59.2$ & $62.8$ \\
\midrule
Average & $23.9$ & $29.1$ & $36.7$ & $63.7$ & $63.0$ & $69.5$ \\
\bottomrule
\end{tabular}
}
\end{adjustbox}
\caption{Task accuracy ($\%$) under $10\%$ noise of SC and the initial rule in Hypothesis Refinement (HR-0) and SRR (SRR-0) on different datasets.}
\label{tab:sc_srr}
\end{table}

\subsection{Method-wise Effectiveness Comparison}
\label{subsec:mec}
Direct Output may limit the model's reasoning ability, as the model must output the rule directly without any intermediate steps. To assess the robustness of different reasoning paradigms, we compare our Sample-steered Rule Refinement (SRR) method with three reasoning methods: (1) Chain of Thought (CoT;~\citealt{wei2022chain}), which decomposes reasoning into step-by-step rationales; (2) Self-Consistency (SC;~\citealt{wang2023selfconsistency}), which aggregates multiple CoT trajectories through majority voting; (3) Self-Refine (SR;~\citealt{madaan2023selfrefine}), which iteratively improves hypotheses using self-generated feedback; and (4) Iterative Hypothesis Refinement (HR;~\citealt{qiu2024phenomenal}), which iteratively generates and ranks multiple hypotheses at each step, then refines the best hypothesis with feedback from incorrect predictions. Table~\ref{tab:methods_res} presents the task accuracy under $10\%$ noise and the deviation from clean data for GPT-4o and DeepSeek-V3.

\paragraph{Superior Performance of SRR} As shown in Table~\ref{tab:methods_res}, our SRR framework achieves state-of-the-art performance across $10$ out of $14$ task-model combinations while exhibiting relatively low performance degradation ($2.1\%$ average gap). This dual advantage stems from two mechanisms: (1) \textit{Diversity-aware hypothesis generation} explores broader solution spaces through subset sampling, outperforming SC's majority voting and the Hypothesis Refinement~\cite{qiu2024phenomenal} that amplifies the similar patterns at the rule initialization stage (Table~\ref{tab:sc_srr}); (2) \textit{Execution-guided feedback} leverages code interpreter for objective error detection, circumventing LLMs' inherent deductive limitations~\cite{chen2023teachinglargelanguagemodels,cheng2024inductivedeductiverethinkingfundamental,zheng2025cursecotlimitationschainofthought}.

\begin{figure}[ht]
    \centering
    \includegraphics[width=0.4\textwidth]{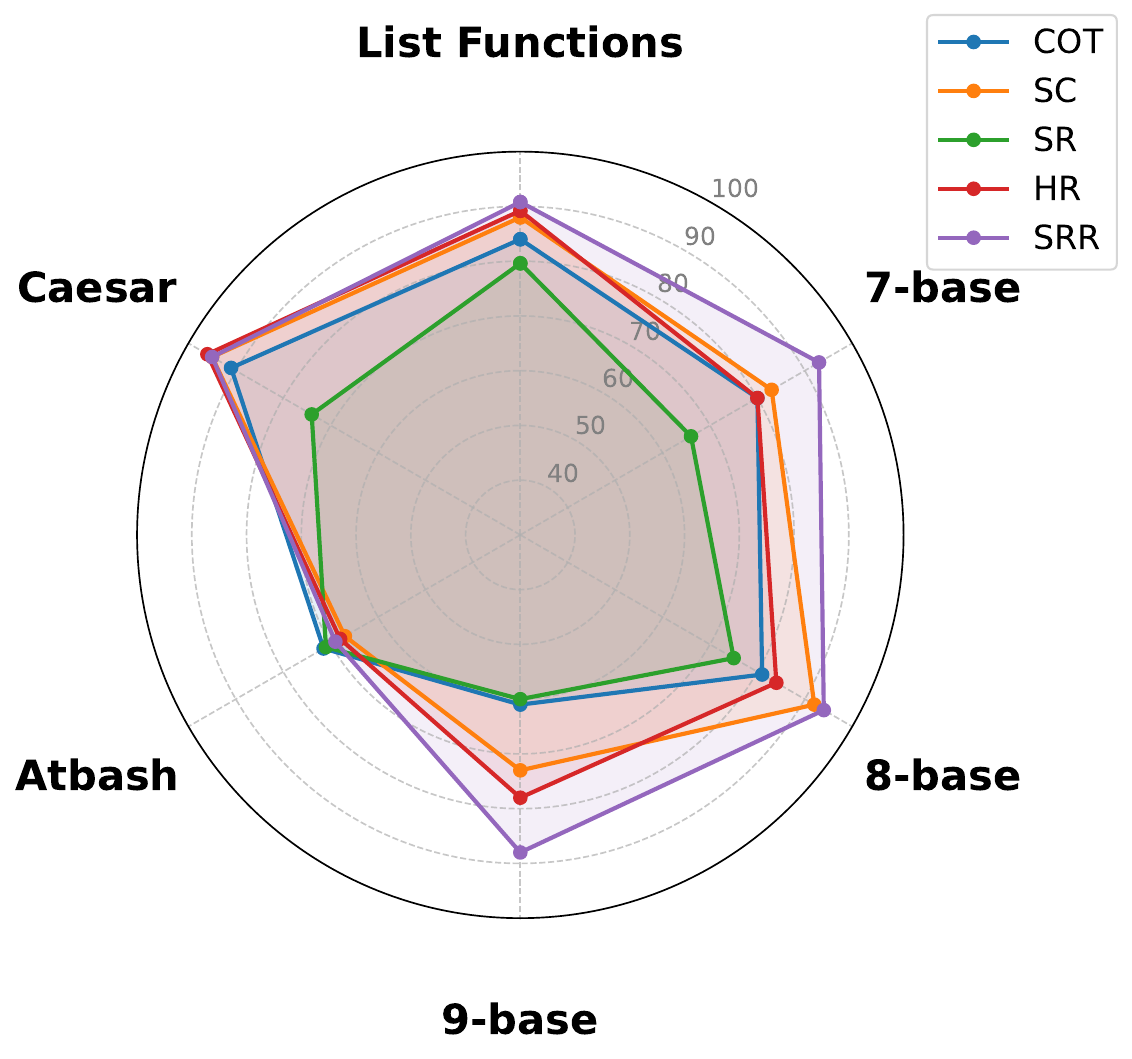}
    \caption{Consistency score $(\%)$ between clean data and data with $10\%$ noise of DeepSeek-V3.}
    \label{fig:consistency_score_method}
\end{figure}

\paragraph{Consistency Scores Reveal Hidden Instability} While task accuracy provides comprehensive insights, consistency scores uncover fundamental reasoning fragility. As depicted in Figure~\ref{fig:consistency_score_method}, the Atbash cipher task exhibits particularly low consistency despite modest accuracy changes. The slight variation in accuracy may merely be an illusion created by the combined effects of noise interference (solved$\rightarrow$unsolved cases) and incorrect generalization (unsolved$\rightarrow$solved cases). Prior and concurrent works' singular focus on accuracy fluctuations~\cite{qiu2024phenomenal,zhou2024can, mirzadeh2024gsmsymbolicunderstandinglimitationsmathematical, huang2025mathperturbbenchmarkingllmsmath} overlooks this critical duality in reasoning robustness.

\paragraph{Counterfactual Challenges Expose Knowledge Reliance} We observe dramatic performance gaps on counterfactual tasks like base-7 and base-9 arithmetic ($90\%+$ vs. $10\%-$ for DeepSeek-V3 vs. GPT-4o) and Keyboard ciphers. Diagnostic analysis of responses from LLMs shows models default to familiar rules rather than induction. GPT-4o persistently misinterprets base-7 and base-9 addition as ``base-8'' or ``decimal sum with constant'', while Atbash and Keyboard ciphers get erroneously classified as Caesar shifts in the reasoning process or self-generated feedback. Although Atbash and Caesar ciphers exhibit similar levels of complexity in transformation, their performance gaps are significant.  
These failures reveal that there is a pattern overfitting in the reasoning process. Current models' inductive reasoning essentially operates through pattern matching rather than abstract induction. When the scarcity of counterfactual tasks in training data forces models to rely on ``real'' rule induction with unseen rules rather than recitation, model performance plummets dramatically.

\subsection{Extended Explorations}
\label{subsec:ees}
To further investigate the robustness of LLMs' inductive reasoning, we conduct a comparative evaluation against DeepSeek-R1~\cite{deepseekai2025deepseekr1incentivizingreasoningcapability}, one of the state-of-the-art reasoning models, and human reasoning patterns.

\paragraph{Comparison with DeepSeek-R1} We focus on the Atbash cipher and List Functions. Atbash provides unified task semantics yet challenges models with unfamiliar transformation logic, while List Functions captures diverse rule abstraction scenarios. Table~\ref{tab:r1_res} compares task accuracy and consistency scores under clean and $10\%$ noise conditions. Notably, DeepSeek-R1 achieves higher task accuracy on both tasks than DeepSeek-V3. However, its consistency scores are still not competitive enough, indicating unresolved instability. Detailed breakdowns of consistency scores are shown in~\ref{tab:r1_stat}. For Atbash, the comparable number of right-to-wrong and wrong-to-right suggests instability and randomness in its reasoning process. Manual inspection shows DeepSeek-R1 still interprets Atbash cipher as character shifts in failed cases, mirroring previous pattern-overfitting behavior. For List Functions, the elevated right-to-wrong rate indicates its sensitivity to input perturbations. 
\begin{table}[ht]
    \centering
    \begin{adjustbox}{max width=1\linewidth}
    {
    \begin{tabular}{c|c|ccc}
    \toprule
        \textbf{Dataset} & \textbf{Method} & \textbf{Clean Acc} & \textbf{Noise Acc} &  \textbf{Consistency} \\
    \midrule
        % \multirow{2}{*}{Atbash} & SRR & $53.0$ & $52.0 (-1.0)$ & $69.0$  \\
        \multirow{2}{*}{Atbash} & CoT & $65.0$ & $65.0 (\pm0.0)$ & $70.0$ \\
        & SRR & $82.0$ & $79.0 (-3.0)$ & $93.0$  \\
    \midrule
        % \multirow{2}{*}{List Func.} & SRR & $67.6$ & $64.8 (-2.8)$ & $90.8$ \\
         \multirow{2}{*}{List Func.} & CoT & $76.0$ & $68.4(-7.6)$ & $87.6$  \\
         & SRR & $83.6$ & $78.0(-5.6)$ & $89.6$  \\
    \bottomrule
    \end{tabular}
    }\end{adjustbox}
    \caption{Task accuracy (\%) and consistency score (\%) between DeepSeek-R1 with CoT and SRR under clean and noisy conditions.}
    \label{tab:r1_res}
\end{table}
\begin{table}[ht]
    \centering
    \begin{adjustbox}{max width=1\linewidth}
    {
    \begin{tabular}{c|c|cccc}
    \toprule
        \textbf{Dataset} & \textbf{Method} & \# BothR & \# BothW &  \# RtoW & \# WtoR \\
    \midrule
        % \multirow{2}{*}{Atbash} & SRR & $37$ & $32$ & $16$ & $15$ \\
        \multirow{2}{*}{Atbash} & CoT & $50$ & $20$ & $15$ & $15$\\
         & SRR & $77$ & $16$ & $5$ & $2$ \\
    \midrule
        % \multirow{2}{*}{List Func.} & SRR & $154$ & $73$ & $15$ & $8$ \\
         \multirow{2}{*}{List Func.} & CoT & $165$ & $54$ & $25$ & $6$\\
         & SRR & $189$ & $35$ & $20$ & $6$ \\
    \bottomrule
    \end{tabular}
    }\end{adjustbox}
    \caption{Detailed breakdowns of consistency score. BothR, BothW, RtoW, WtoR represent both right, both wrong, right to wrong from clean condition to noisy condition and wrong to right respectively.}
    \label{tab:r1_stat}
\end{table}

\begin{figure}[ht]
    \centering
    \includegraphics[width=0.4\textwidth]{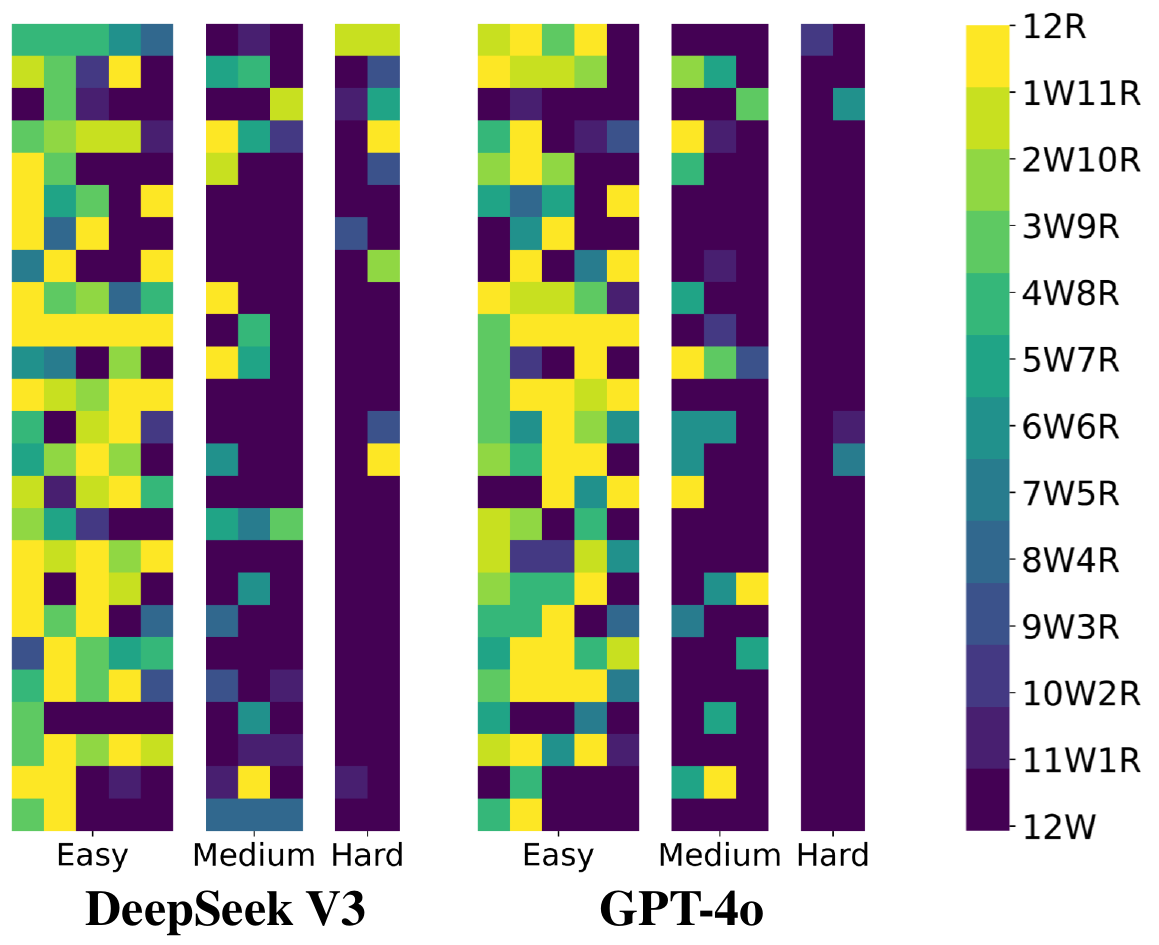}
    \caption{Task-solving consistency of DeepSeek-V3 and GPT-4o on List Functions. Each cell represents a task, arranged by ascending difficulty (top-to-bottom, left-to-right). Colors denote correctness patterns: 12R (all correct) to 12W (all wrong), with intermediate states (e.g., 1W11R: 1 wrong, 11 correct).}
    \label{fig:consistency_score_lf}
\end{figure}

\paragraph{Human Reasoning Comparison}~\citet{rule2020child} reports human performance on List Functions. We divide the tasks into three difficulty levels in a $5:3:2$ ratio based on the sorted mean human performance. To contrast with human-like thinking patterns, we analyze consistency across 12 trials (4 noise levels $\times$ 3 runs) in Section~\ref{subsec:noise_level}. Figure~\ref{fig:consistency_score_lf} visualizes the distribution of consistency and performance per task in List Functions, stratified by difficulty. LLMs show macro-level alignment yet micro-level divergence compared to human. While LLMs broadly mirror human-like stability, showing higher consistency on both simple and hard tasks, with moderate instability on medium-difficulty tasks, their internal patterns reveal critical deviations. For hard tasks, models display unpredictability (e.g., sporadic success) rather than systematic incapacity. Simple tasks, despite their low complexity, exhibit in consistent performance unrelated to intrinsic difficulty. Notably, GPT-4o and DeepSeek-V3 demonstrate similar behavioral trends, suggesting a shared inductive bias.

\section{Discussion}
\label{subsec:ana}
Our experimental results reveal limitations in LLMs' inductive reasoning. We try to analyze the results in this section. 

\paragraph{Process Analysis} Inductive reasoning can be conceptualized through a Bayesian paradigm~\cite{doi:10.1126/science.1192788}, where models update posterior distributions over a hypothesis space based on observations. The introduction of noise disrupts the reasoning through dual mechanisms: \textbf{Noise-Induced Hypothesis Drift}, where the initial hypothesis space becomes misaligned with true rules when noise introduces conflicting patterns. This drift particularly impacts Direct Output (DO) method, which lacks intermediate reasoning steps and relies heavily on the initial hypothesis space. This is more evident for tasks with similar internal patterns, the model's performance fluctuates significantly (Cryptography) or even fails (Arithmetic) compared to other methods. 
\textbf{Evidence Ambiguity Amplification}, where noise reduces the effective signal-to-noise ratio during posterior optimization. Methods with iterative or step-by-step reasoning suffer error accumulation cascades\textemdash each reasoning step propagates uncertainty. As shown in Table~\ref{tab:methods_res}, CoT and SR exhibit larger average accuracy drops and worse performance. SRR avoids it by objectively identifying errors through execution. 

\paragraph{Implications for Robust Inductive Reasoning}
LLMs heavily rely on prior knowledge rather than pure induction. They blend memorized pattern matching with shallow reasoning. When observations align with priors (e.g., Caesar cipher), they demonstrate pseudo-robustness. However, conflicting patterns and counterfactual scenarios expose this fragility\textemdash models either default to familiar templates or enter unstable hypothesis oscillations, as shown in Table~\ref{tab:case_study}. Even the state-of-the-art reasoning model exhibits unstable performance in unified tasks. This contrasts sharply with human few-shot learning, where true rule abstraction enables stable generalization~\cite{doi:10.1126/science.aab3050}. Achieving human-level robust reasoning and few-shot induction requires disentangling knowledge recitation from inductive rule formation.

\begin{table}[ht]
    \centering
    \small
    \begin{tabularx}{\linewidth}{X}
    \toprule
    \rowcolor[gray]{0.9}\textbf{Unstable Hypothesis Oscillation}\\
    \midrule
   \textbf{Seen Examples} (base-9 arithmetic):
    I: 68+68 O: 147; \\
    I: 80+42 O: 132; I: 76+17 O: 104; \textcolor{red}{I: 57+56 O: 113};\\
    I: 38+14 O: 53; I: 43+84 O: 137; I: 48+55 O: 114; ...\\
    \midrule
    \textbf{Answer}: To solve this problem, let's analyze the examples to identify a potential pattern: 1. 68 + 68 = 147: The expected output of 68 + 68 should be 136, but the output is 147. This suggests there is a transformation applied to the arithmetic result. 2. 80 + 42 = 132: Normally, 80 + 42 equals 122, but the output is 132. ... 4. 57 + 56 = 113: The expected result is 113, which matches the output. ... \textcolor{orange}{The examples seem to suggest a pattern of adding an extra 10 to the arithmetic sum of the inputs, except in the case where the sum already matches the desired output (as in the case of 57 + 56 = 113). With these observations, we can propose a rule: If the sum of the numbers is not equal to the expected output, add 10 to the arithmetic sum.}\\
    \midrule
    \rowcolor[gray]{0.9}\textbf{Default to Familiar Templates}\\
    \midrule
    \textbf{Seen Examples} (Atbach cipher): I: cuissard O: xfrhhziw; \\
    I: tripsis", O: girkhrh; I: Maccabaeus O: Nzxxzyzvfh; ...\\
    \midrule
    \textbf{Answer:} To solve this problem, let's analyze the given examples to find a pattern: 1.  cuissard->xfrhhziw 2. tripsis->girkhrh 3. Maccabaeus->Nzxxzyzvfh ... Observations: 1. Character Shifts: The transformation seems to involve shifting characters. For example, `c' in ``cuissar'' becomes `x' in ``xfrhhziw''. \textcolor{orange}{This suggests a possible Caesar cipher or similar character shift.} ... 4.  Pattern Recognition: The shifts are not consistent across all examples, suggesting that some examples might be incorrect or that multiple rules are being applied. However, a common shift that appears frequently is 21 positions backward (e.g., `c' to 'x', `P' to `K', `C' to `X', `W' to `D'). Hypothesis: \textcolor{orange}{Given the frequent appearance of a 21-position backward shift, let's hypothesize that the rule involves shifting each character 21 positions backward in the alphabet.}\\
    \bottomrule
    \end{tabularx}
    \caption{The failure cases of the models on the Arithmetic and Cryptography tasks. The \textcolor{red}{red} text indicates the noise introduced, and the \textcolor{orange}{orange} text indicates the wrong hypothesis of the model.}
    \label{tab:case_study}
\end{table}

\section{Related Work}

\paragraph{Inductive Reasoning} Early work~\cite{Heit2000} established foundational properties of inductive reasoning. In cognitive science, induction is considered a process of probabilistic belief updating within the Bayesian paradigm~\cite{doi:10.1126/science.1192788}. Human cognition and learning are explained through the integration of prior knowledge with observed data to compute posterior distributions. Comparative studies~\cite{doi:10.1126/science.aab3050,Lake_Ullman_Tenenbaum_Gershman_2017} have highlighted the contrast between human learners and machine intelligence.
With the advent of pre-trained language models~\cite{brown2020language}, research on inductive reasoning has shifted from domain-specific and neural formulation~\cite{tian2020learning,odena2021bustlebottomupprogramsynthesis,SABLEMEYER2022101527} to natural language. Initial approaches~\cite{pmlr-v139-alet21a,ijcai2024p693,mirchandani2023large,yang-etal-2024-language} predominantly relied on input-output (IO) prompting, which evaluates model performance on unseen examples without explicit rule articulation. However, this paradigm overlooks the rule-inference process, as it conflates rule induction with rule execution capabilities. Recent efforts~\cite{wang2024hypothesis,qiu2024phenomenal} explicitly evaluate intermediate rules, and our settings and methods are similar to theirs, despite the sampling part in initialization and iteration. Specifically, ~\citet{qiu2024phenomenal} also proposed an evaluation on noisy conditions, but their analysis is confined to a single dataset and prioritized accuracy metrics. We further explore the consistency and draw a different conclusion that LLMs' inductive reasoning capabilities are not robust and generalizable.

\paragraph{Robustness of Reasoning in LLMs}
In this work, robustness refers to the ability to maintain consistent performance under imperfections or counterfactual scenarios~\cite{elazar-etal-2021-measuring}, which is intrinsically linked to the diversity and unpredictability of the generation process~\cite{zhang2023sirenssongaiocean,huang2025hallu}. Prior works on LLM behavioral variability exhibit inconsistent performance across time~\cite{tu2024chatlogcarefullyevaluatingevolution,chen2023chatgptsbehaviorchangingtime}. Recent works focus on understanding their internal reasoning mechanisms by altering conditions, such as changing numerical values in mathematical tasks~\cite{mirzadeh2024gsmsymbolicunderstandinglimitationsmathematical,huang2025mathperturbbenchmarkingllmsmath} or examining performance on counterfactual tasks~\cite{wu-etal-2024-reasoning}. \citet{zhou2024can} investigates the impact of noisy rationales on model performance. These studies primarily measure robustness through global accuracy changes, overlooking inconsistencies at the instance level. Our work addresses this gap by introducing a consistency score that quantifies intra-task stability, providing a more granular view of model robustness.

\section{Conclusion}
In this paper, we explore the robustness of inductive reasoning in large language models under imperfect observations. Through introducing the Robust Rule Induction task with metrics of both holistic and individual levels, we systematically evaluate the ability of LLMs to maintain stable and consistent rule abstraction. The Sample-steered Rule Refinement outperforms other reasoning paradigms by effectively leveraging diversity-aware hypothesis generation and execution-guided feedback. Our findings reveal that while LLMs can demonstrate impressive reasoning capabilities, they are inherently sensitive to noise and prone to hypothesis drifting and pattern overfitting.

\section*{Limitations}
We discuss the limitations of our work as follows. First, the general goal of our research is to investigate the relationship between a model's reasoning and its understanding, specifically examining whether correctly solving a problem indicates true comprehension and reasoning of the underlying issue. To this end, our evaluation primarily focuses on highly formalized and symbolic tasks that are easily verifiable. However, real-world inductive reasoning often involves ambiguous rules like social norms from text or visual patterns, which are not captured by our formalism. Second, the tasks in Arithmetic and Cryptography may be relatively simple and unified, potentially limiting insights into the capabilities of large language models. Expanding task diversity could uncover deeper limitations in LLMs' reasoning abilities. Finally, conducting a large-scale assessment of human performance across all tasks under varying levels of noise was beyond the scope of this study, which restricts our ability to compare human and model reasoning patterns comprehensively. 

\section*{Ethics Statement}
We discuss the ethical considerations of our work in the context of the following aspects: (1) \textbf{Data collection and use}. We use publicly available datasets and self-generated data for evaluation. We ensure that the data is only used for academic research purposes and no personal data is involved. The List Functions dataset is released with Apache-2.0 license\footnote{\url{https://www.apache.org/licenses/LICENSE-2.0}} in Big-bench~\cite{srivastava2023beyond}. We strictly follow the license terms and conditions. (2) \textbf{LLMs API}. We comply with the terms of service of the LLMs API providers strictly, maintaining fair use. (3) \textbf{Transparency and reproducibility}. We provide detailed descriptions of our methods, datasets, and evaluation metrics to ensure transparency. Our code and evaluation results are publicly available. (4) \textbf{AI assistance}. We use ChatGPT to paraphrase some sentences.

\section*{Acknowledgement}
We thank all the anonymous reviewers and meta reviewers for their valuable comments. The authors of this paper were supported by the ITSP Platform Research Project (ITS/189/23FP) from ITC of Hong Kong, SAR, China, and the AoE (AoE/E-601/24-N), the RIF (R6021-20) and the GRF (16205322) from RGC of Hong Kong, SAR, China.

\bibliography{custom}

\begin{thebibliography}{43}
\providecommand{\natexlab}[1]{#1}

\bibitem[{Alet et~al.(2021)Alet, Lopez-Contreras, Koppel, Nye, Solar-Lezama, Lozano-Perez, Kaelbling, and Tenenbaum}]{pmlr-v139-alet21a}
Ferran Alet, Javier Lopez-Contreras, James Koppel, Maxwell Nye, Armando Solar-Lezama, Tomas Lozano-Perez, Leslie Kaelbling, and Joshua Tenenbaum. 2021.
\newblock \href {https://proceedings.mlr.press/v139/alet21a.html} {A large-scale benchmark for few-shot program induction and synthesis}.
\newblock In \emph{Proceedings of the 38th International Conference on Machine Learning}, volume 139 of \emph{Proceedings of Machine Learning Research}, pages 175--186. PMLR.

\bibitem[{bench authors(2023)}]{srivastava2023beyond}
BIG bench authors. 2023.
\newblock \href {https://openreview.net/forum?id=uyTL5Bvosj} {Beyond the imitation game: Quantifying and extrapolating the capabilities of language models}.
\newblock \emph{Transactions on Machine Learning Research}.

\bibitem[{Brown et~al.(2020)Brown, Mann, Ryder, Subbiah, Kaplan, Dhariwal, Neelakantan, Shyam, Sastry, Askell, Agarwal, Herbert-Voss, Krueger, Henighan, Child, Ramesh, Ziegler, Wu, Winter, Hesse, Chen, Sigler, Litwin, Gray, Chess, Clark, Berner, McCandlish, Radford, Sutskever, and Amodei}]{brown2020language}
Tom~B. Brown, Benjamin Mann, Nick Ryder, Melanie Subbiah, Jared Kaplan, Prafulla Dhariwal, Arvind Neelakantan, Pranav Shyam, Girish Sastry, Amanda Askell, Sandhini Agarwal, Ariel Herbert-Voss, Gretchen Krueger, Tom Henighan, Rewon Child, Aditya Ramesh, Daniel~M. Ziegler, Jeffrey Wu, Clemens Winter, Christopher Hesse, Mark Chen, Eric Sigler, Mateusz Litwin, Scott Gray, Benjamin Chess, Jack Clark, Christopher Berner, Sam McCandlish, Alec Radford, Ilya Sutskever, and Dario Amodei. 2020.
\newblock Language models are few-shot learners.
\newblock In \emph{Proceedings of the 34th International Conference on Neural Information Processing Systems}, NIPS '20, Red Hook, NY, USA. Curran Associates Inc.

\bibitem[{Chen et~al.(2023{\natexlab{a}})Chen, Zaharia, and Zou}]{chen2023chatgptsbehaviorchangingtime}
Lingjiao Chen, Matei Zaharia, and James Zou. 2023{\natexlab{a}}.
\newblock \href {https://arxiv.org/abs/2307.09009} {How is chatgpt's behavior changing over time?}
\newblock \emph{Preprint}, arXiv:2307.09009.

\bibitem[{Chen et~al.(2023{\natexlab{b}})Chen, Lin, Schärli, and Zhou}]{chen2023teachinglargelanguagemodels}
Xinyun Chen, Maxwell Lin, Nathanael Schärli, and Denny Zhou. 2023{\natexlab{b}}.
\newblock \href {https://arxiv.org/abs/2304.05128} {Teaching large language models to self-debug}.
\newblock \emph{Preprint}, arXiv:2304.05128.

\bibitem[{Cheng et~al.(2024)Cheng, Yang, Jiang, Wang, Huang, Li, Li, Li, Gao, Li, Yin, and Sun}]{cheng2024inductivedeductiverethinkingfundamental}
Kewei Cheng, Jingfeng Yang, Haoming Jiang, Zhengyang Wang, Binxuan Huang, Ruirui Li, Shiyang Li, Zheng Li, Yifan Gao, Xian Li, Bing Yin, and Yizhou Sun. 2024.
\newblock \href {https://arxiv.org/abs/2408.00114} {Inductive or deductive? rethinking the fundamental reasoning abilities of llms}.
\newblock \emph{Preprint}, arXiv:2408.00114.

\bibitem[{Chollet(2019)}]{chollet2019measureintelligence}
François Chollet. 2019.
\newblock \href {https://arxiv.org/abs/1911.01547} {On the measure of intelligence}.
\newblock \emph{Preprint}, arXiv:1911.01547.

\bibitem[{Collins et~al.(2024)Collins, Sucholutsky, Bhatt, Chandra, Wong, Lee, Zhang, Zhi-Xuan, Ho, Mansinghka, Weller, Tenenbaum, and Griffiths}]{collins_building_2024}
Katherine~M. Collins, Ilia Sucholutsky, Umang Bhatt, Kartik Chandra, Lionel Wong, Mina Lee, Cedegao~E. Zhang, Tan Zhi-Xuan, Mark Ho, Vikash Mansinghka, Adrian Weller, Joshua~B. Tenenbaum, and Thomas~L. Griffiths. 2024.
\newblock \href {https://doi.org/10.1038/s41562-024-01991-9} {Building machines that learn and think with people}.
\newblock \emph{Nature Human Behaviour}, 8(10):1851--1863.
\newblock Publisher: Nature Publishing Group.

\bibitem[{DeepSeek-AI et~al.(2025)DeepSeek-AI, Guo, Yang, Zhang, Song, Zhang, Xu, Zhu, Ma, Wang, Bi, Zhang, Yu, Wu, Wu, Gou, Shao, Li, Gao, Liu, Xue, Wang, Wu, Feng, Lu, Zhao, Deng, Zhang, Ruan, Dai, Chen, Ji, Li, Lin, Dai, Luo, Hao, Chen, Li, Zhang, Bao, Xu, Wang, Ding, Xin, Gao, Qu, Li, Guo, Li, Wang, Chen, Yuan, Qiu, Li, Cai, Ni, Liang, Chen, Dong, Hu, Gao, Guan, Huang, Yu, Wang, Zhang, Zhao, Wang, Zhang, Xu, Xia, Zhang, Zhang, Tang, Li, Wang, Li, Tian, Huang, Zhang, Wang, Chen, Du, Ge, Zhang, Pan, Wang, Chen, Jin, Chen, Lu, Zhou, Chen, Ye, Wang, Yu, Zhou, Pan, Li, Zhou, Wu, Ye, Yun, Pei, Sun, Wang, Zeng, Zhao, Liu, Liang, Gao, Yu, Zhang, Xiao, An, Liu, Wang, Chen, Nie, Cheng, Liu, Xie, Liu, Yang, Li, Su, Lin, Li, Jin, Shen, Chen, Sun, Wang, Song, Zhou, Wang, Shan, Li, Wang, Wei, Zhang, Xu, Li, Zhao, Sun, Wang, Yu, Zhang, Shi, Xiong, He, Piao, Wang, Tan, Ma, Liu, Guo, Ou, Wang, Gong, Zou, He, Xiong, Luo, You, Liu, Zhou, Zhu, Xu, Huang, Li, Zheng, Zhu, Ma, Tang, Zha, Yan, Ren, Ren, Sha, Fu, Xu, Xie, Zhang,
  Hao, Ma, Yan, Wu, Gu, Zhu, Liu, Li, Xie, Song, Pan, Huang, Xu, Zhang, and Zhang}]{deepseekai2025deepseekr1incentivizingreasoningcapability}
DeepSeek-AI, Daya Guo, Dejian Yang, Haowei Zhang, Junxiao Song, Ruoyu Zhang, Runxin Xu, Qihao Zhu, Shirong Ma, Peiyi Wang, Xiao Bi, Xiaokang Zhang, Xingkai Yu, Yu~Wu, Z.~F. Wu, Zhibin Gou, Zhihong Shao, Zhuoshu Li, Ziyi Gao, Aixin Liu, Bing Xue, Bingxuan Wang, Bochao Wu, Bei Feng, Chengda Lu, Chenggang Zhao, Chengqi Deng, Chenyu Zhang, Chong Ruan, Damai Dai, Deli Chen, Dongjie Ji, Erhang Li, Fangyun Lin, Fucong Dai, Fuli Luo, Guangbo Hao, Guanting Chen, Guowei Li, H.~Zhang, Han Bao, Hanwei Xu, Haocheng Wang, Honghui Ding, Huajian Xin, Huazuo Gao, Hui Qu, Hui Li, Jianzhong Guo, Jiashi Li, Jiawei Wang, Jingchang Chen, Jingyang Yuan, Junjie Qiu, Junlong Li, J.~L. Cai, Jiaqi Ni, Jian Liang, Jin Chen, Kai Dong, Kai Hu, Kaige Gao, Kang Guan, Kexin Huang, Kuai Yu, Lean Wang, Lecong Zhang, Liang Zhao, Litong Wang, Liyue Zhang, Lei Xu, Leyi Xia, Mingchuan Zhang, Minghua Zhang, Minghui Tang, Meng Li, Miaojun Wang, Mingming Li, Ning Tian, Panpan Huang, Peng Zhang, Qiancheng Wang, Qinyu Chen, Qiushi Du, Ruiqi Ge, Ruisong
  Zhang, Ruizhe Pan, Runji Wang, R.~J. Chen, R.~L. Jin, Ruyi Chen, Shanghao Lu, Shangyan Zhou, Shanhuang Chen, Shengfeng Ye, Shiyu Wang, Shuiping Yu, Shunfeng Zhou, Shuting Pan, S.~S. Li, Shuang Zhou, Shaoqing Wu, Shengfeng Ye, Tao Yun, Tian Pei, Tianyu Sun, T.~Wang, Wangding Zeng, Wanjia Zhao, Wen Liu, Wenfeng Liang, Wenjun Gao, Wenqin Yu, Wentao Zhang, W.~L. Xiao, Wei An, Xiaodong Liu, Xiaohan Wang, Xiaokang Chen, Xiaotao Nie, Xin Cheng, Xin Liu, Xin Xie, Xingchao Liu, Xinyu Yang, Xinyuan Li, Xuecheng Su, Xuheng Lin, X.~Q. Li, Xiangyue Jin, Xiaojin Shen, Xiaosha Chen, Xiaowen Sun, Xiaoxiang Wang, Xinnan Song, Xinyi Zhou, Xianzu Wang, Xinxia Shan, Y.~K. Li, Y.~Q. Wang, Y.~X. Wei, Yang Zhang, Yanhong Xu, Yao Li, Yao Zhao, Yaofeng Sun, Yaohui Wang, Yi~Yu, Yichao Zhang, Yifan Shi, Yiliang Xiong, Ying He, Yishi Piao, Yisong Wang, Yixuan Tan, Yiyang Ma, Yiyuan Liu, Yongqiang Guo, Yuan Ou, Yuduan Wang, Yue Gong, Yuheng Zou, Yujia He, Yunfan Xiong, Yuxiang Luo, Yuxiang You, Yuxuan Liu, Yuyang Zhou, Y.~X. Zhu,
  Yanhong Xu, Yanping Huang, Yaohui Li, Yi~Zheng, Yuchen Zhu, Yunxian Ma, Ying Tang, Yukun Zha, Yuting Yan, Z.~Z. Ren, Zehui Ren, Zhangli Sha, Zhe Fu, Zhean Xu, Zhenda Xie, Zhengyan Zhang, Zhewen Hao, Zhicheng Ma, Zhigang Yan, Zhiyu Wu, Zihui Gu, Zijia Zhu, Zijun Liu, Zilin Li, Ziwei Xie, Ziyang Song, Zizheng Pan, Zhen Huang, Zhipeng Xu, Zhongyu Zhang, and Zhen Zhang. 2025.
\newblock \href {https://arxiv.org/abs/2501.12948} {Deepseek-r1: Incentivizing reasoning capability in llms via reinforcement learning}.
\newblock \emph{Preprint}, arXiv:2501.12948.

\bibitem[{DeepSeek-AI et~al.(2024)DeepSeek-AI, Liu, Feng, Xue, Wang, Wu, Lu, Zhao, Deng, Zhang, Ruan, Dai, Guo, Yang, Chen, Ji, Li, Lin, Dai, Luo, Hao, Chen, Li, Zhang, Bao, Xu, Wang, Zhang, Ding, Xin, Gao, Li, Qu, Cai, Liang, Guo, Ni, Li, Wang, Chen, Chen, Yuan, Qiu, Li, Song, Dong, Hu, Gao, Guan, Huang, Yu, Wang, Zhang, Xu, Xia, Zhao, Wang, Zhang, Li, Wang, Zhang, Zhang, Tang, Li, Tian, Huang, Wang, Zhang, Wang, Zhu, Chen, Du, Chen, Jin, Ge, Zhang, Pan, Wang, Xu, Zhang, Chen, Li, Lu, Zhou, Chen, Wu, Ye, Ye, Ma, Wang, Zhou, Yu, Zhou, Pan, Wang, Yun, Pei, Sun, Xiao, Zeng, Zhao, An, Liu, Liang, Gao, Yu, Zhang, Li, Jin, Wang, Bi, Liu, Wang, Shen, Chen, Zhang, Chen, Nie, Sun, Wang, Cheng, Liu, Xie, Liu, Yu, Song, Shan, Zhou, Yang, Li, Su, Lin, Li, Wang, Wei, Zhu, Zhang, Xu, Xu, Huang, Li, Zhao, Sun, Li, Wang, Yu, Zheng, Zhang, Shi, Xiong, He, Tang, Piao, Wang, Tan, Ma, Liu, Guo, Wu, Ou, Zhu, Wang, Gong, Zou, He, Zha, Xiong, Ma, Yan, Luo, You, Liu, Zhou, Wu, Ren, Ren, Sha, Fu, Xu, Huang, Zhang, Xie, Zhang, Hao,
  Gou, Ma, Yan, Shao, Xu, Wu, Zhang, Li, Gu, Zhu, Liu, Li, Xie, Song, Gao, and Pan}]{deepseekai2024deepseekv3technicalreport}
DeepSeek-AI, Aixin Liu, Bei Feng, Bing Xue, Bingxuan Wang, Bochao Wu, Chengda Lu, Chenggang Zhao, Chengqi Deng, Chenyu Zhang, Chong Ruan, Damai Dai, Daya Guo, Dejian Yang, Deli Chen, Dongjie Ji, Erhang Li, Fangyun Lin, Fucong Dai, Fuli Luo, Guangbo Hao, Guanting Chen, Guowei Li, H.~Zhang, Han Bao, Hanwei Xu, Haocheng Wang, Haowei Zhang, Honghui Ding, Huajian Xin, Huazuo Gao, Hui Li, Hui Qu, J.~L. Cai, Jian Liang, Jianzhong Guo, Jiaqi Ni, Jiashi Li, Jiawei Wang, Jin Chen, Jingchang Chen, Jingyang Yuan, Junjie Qiu, Junlong Li, Junxiao Song, Kai Dong, Kai Hu, Kaige Gao, Kang Guan, Kexin Huang, Kuai Yu, Lean Wang, Lecong Zhang, Lei Xu, Leyi Xia, Liang Zhao, Litong Wang, Liyue Zhang, Meng Li, Miaojun Wang, Mingchuan Zhang, Minghua Zhang, Minghui Tang, Mingming Li, Ning Tian, Panpan Huang, Peiyi Wang, Peng Zhang, Qiancheng Wang, Qihao Zhu, Qinyu Chen, Qiushi Du, R.~J. Chen, R.~L. Jin, Ruiqi Ge, Ruisong Zhang, Ruizhe Pan, Runji Wang, Runxin Xu, Ruoyu Zhang, Ruyi Chen, S.~S. Li, Shanghao Lu, Shangyan Zhou, Shanhuang
  Chen, Shaoqing Wu, Shengfeng Ye, Shengfeng Ye, Shirong Ma, Shiyu Wang, Shuang Zhou, Shuiping Yu, Shunfeng Zhou, Shuting Pan, T.~Wang, Tao Yun, Tian Pei, Tianyu Sun, W.~L. Xiao, Wangding Zeng, Wanjia Zhao, Wei An, Wen Liu, Wenfeng Liang, Wenjun Gao, Wenqin Yu, Wentao Zhang, X.~Q. Li, Xiangyue Jin, Xianzu Wang, Xiao Bi, Xiaodong Liu, Xiaohan Wang, Xiaojin Shen, Xiaokang Chen, Xiaokang Zhang, Xiaosha Chen, Xiaotao Nie, Xiaowen Sun, Xiaoxiang Wang, Xin Cheng, Xin Liu, Xin Xie, Xingchao Liu, Xingkai Yu, Xinnan Song, Xinxia Shan, Xinyi Zhou, Xinyu Yang, Xinyuan Li, Xuecheng Su, Xuheng Lin, Y.~K. Li, Y.~Q. Wang, Y.~X. Wei, Y.~X. Zhu, Yang Zhang, Yanhong Xu, Yanhong Xu, Yanping Huang, Yao Li, Yao Zhao, Yaofeng Sun, Yaohui Li, Yaohui Wang, Yi~Yu, Yi~Zheng, Yichao Zhang, Yifan Shi, Yiliang Xiong, Ying He, Ying Tang, Yishi Piao, Yisong Wang, Yixuan Tan, Yiyang Ma, Yiyuan Liu, Yongqiang Guo, Yu~Wu, Yuan Ou, Yuchen Zhu, Yuduan Wang, Yue Gong, Yuheng Zou, Yujia He, Yukun Zha, Yunfan Xiong, Yunxian Ma, Yuting Yan, Yuxiang
  Luo, Yuxiang You, Yuxuan Liu, Yuyang Zhou, Z.~F. Wu, Z.~Z. Ren, Zehui Ren, Zhangli Sha, Zhe Fu, Zhean Xu, Zhen Huang, Zhen Zhang, Zhenda Xie, Zhengyan Zhang, Zhewen Hao, Zhibin Gou, Zhicheng Ma, Zhigang Yan, Zhihong Shao, Zhipeng Xu, Zhiyu Wu, Zhongyu Zhang, Zhuoshu Li, Zihui Gu, Zijia Zhu, Zijun Liu, Zilin Li, Ziwei Xie, Ziyang Song, Ziyi Gao, and Zizheng Pan. 2024.
\newblock \href {https://arxiv.org/abs/2412.19437} {Deepseek-v3 technical report}.
\newblock \emph{Preprint}, arXiv:2412.19437.

\bibitem[{Elazar et~al.(2021)Elazar, Kassner, Ravfogel, Ravichander, Hovy, Sch{\"u}tze, and Goldberg}]{elazar-etal-2021-measuring}
Yanai Elazar, Nora Kassner, Shauli Ravfogel, Abhilasha Ravichander, Eduard Hovy, Hinrich Sch{\"u}tze, and Yoav Goldberg. 2021.
\newblock \href {https://doi.org/10.1162/tacl_a_00410} {Measuring and improving consistency in pretrained language models}.
\newblock \emph{Transactions of the Association for Computational Linguistics}, 9:1012--1031.

\bibitem[{Feldman(1997)}]{Feldman1997TheSO}
Jacob Feldman. 1997.
\newblock \href {https://api.semanticscholar.org/CorpusID:15319518} {The structure of perceptual categories}.
\newblock \emph{Journal of mathematical psychology}, 41 2:145--70.

\bibitem[{Gendron et~al.(2024)Gendron, Bao, Witbrock, and Dobbie}]{ijcai2024p693}
Gaël Gendron, Qiming Bao, Michael Witbrock, and Gillian Dobbie. 2024.
\newblock \href {https://doi.org/10.24963/ijcai.2024/693} {Large language models are not strong abstract reasoners}.
\newblock In \emph{Proceedings of the Thirty-Third International Joint Conference on Artificial Intelligence, {IJCAI-24}}, pages 6270--6278. International Joint Conferences on Artificial Intelligence Organization.
\newblock Main Track.

\bibitem[{Heit(2000)}]{Heit2000}
Evan Heit. 2000.
\newblock \href {https://doi.org/10.3758/BF03212996} {Properties of inductive reasoning}.
\newblock \emph{Psychonomic Bulletin {\&} Review}, 7(4):569--592.

\bibitem[{Huang et~al.(2025{\natexlab{a}})Huang, Guo, Li, Ji, Ge, Li, Guo, Cai, Yuan, Wang, Wu, Yin, Tang, Huang, Jin, Chen, Zhang, and Wang}]{huang2025mathperturbbenchmarkingllmsmath}
Kaixuan Huang, Jiacheng Guo, Zihao Li, Xiang Ji, Jiawei Ge, Wenzhe Li, Yingqing Guo, Tianle Cai, Hui Yuan, Runzhe Wang, Yue Wu, Ming Yin, Shange Tang, Yangsibo Huang, Chi Jin, Xinyun Chen, Chiyuan Zhang, and Mengdi Wang. 2025{\natexlab{a}}.
\newblock \href {https://arxiv.org/abs/2502.06453} {Math-perturb: Benchmarking llms' math reasoning abilities against hard perturbations}.
\newblock \emph{Preprint}, arXiv:2502.06453.

\bibitem[{Huang et~al.(2025{\natexlab{b}})Huang, Yu, Ma, Zhong, Feng, Wang, Chen, Peng, Feng, Qin, and Liu}]{huang2025hallu}
Lei Huang, Weijiang Yu, Weitao Ma, Weihong Zhong, Zhangyin Feng, Haotian Wang, Qianglong Chen, Weihua Peng, Xiaocheng Feng, Bing Qin, and Ting Liu. 2025{\natexlab{b}}.
\newblock \href {https://doi.org/10.1145/3703155} {A survey on hallucination in large language models: Principles, taxonomy, challenges, and open questions}.
\newblock \emph{ACM Trans. Inf. Syst.}, 43(2).

\bibitem[{Lake and Baroni(2018)}]{pmlr-v80-lake18a}
Brenden Lake and Marco Baroni. 2018.
\newblock \href {https://proceedings.mlr.press/v80/lake18a.html} {Generalization without systematicity: On the compositional skills of sequence-to-sequence recurrent networks}.
\newblock In \emph{Proceedings of the 35th International Conference on Machine Learning}, volume~80 of \emph{Proceedings of Machine Learning Research}, pages 2873--2882. PMLR.

\bibitem[{Lake et~al.(2015)Lake, Salakhutdinov, and Tenenbaum}]{doi:10.1126/science.aab3050}
Brenden~M. Lake, Ruslan Salakhutdinov, and Joshua~B. Tenenbaum. 2015.
\newblock \href {https://doi.org/10.1126/science.aab3050} {Human-level concept learning through probabilistic program induction}.
\newblock \emph{Science}, 350(6266):1332--1338.

\bibitem[{Lake et~al.(2017)Lake, Ullman, Tenenbaum, and Gershman}]{Lake_Ullman_Tenenbaum_Gershman_2017}
Brenden~M. Lake, Tomer~D. Ullman, Joshua~B. Tenenbaum, and Samuel~J. Gershman. 2017.
\newblock \href {https://doi.org/10.1017/S0140525X16001837} {Building machines that learn and think like people}.
\newblock \emph{Behavioral and Brain Sciences}, 40:e253.

\bibitem[{Liu et~al.(2024)Liu, Neubig, and Andreas}]{liu2024an}
Emmy Liu, Graham Neubig, and Jacob Andreas. 2024.
\newblock \href {https://openreview.net/forum?id=nUNbjMDBWC} {An incomplete loop: Instruction inference, instruction following, and in-context learning in language models}.
\newblock In \emph{First Conference on Language Modeling}.

\bibitem[{Lu et~al.(2022)Lu, Bartolo, Moore, Riedel, and Stenetorp}]{lu-etal-2022-fantastically}
Yao Lu, Max Bartolo, Alastair Moore, Sebastian Riedel, and Pontus Stenetorp. 2022.
\newblock \href {https://doi.org/10.18653/v1/2022.acl-long.556} {Fantastically ordered prompts and where to find them: Overcoming few-shot prompt order sensitivity}.
\newblock In \emph{Proceedings of the 60th Annual Meeting of the Association for Computational Linguistics (Volume 1: Long Papers)}, pages 8086--8098, Dublin, Ireland. Association for Computational Linguistics.

\bibitem[{Madaan et~al.(2023)Madaan, Tandon, Gupta, Hallinan, Gao, Wiegreffe, Alon, Dziri, Prabhumoye, Yang, Gupta, Majumder, Hermann, Welleck, Yazdanbakhsh, and Clark}]{madaan2023selfrefine}
Aman Madaan, Niket Tandon, Prakhar Gupta, Skyler Hallinan, Luyu Gao, Sarah Wiegreffe, Uri Alon, Nouha Dziri, Shrimai Prabhumoye, Yiming Yang, Shashank Gupta, Bodhisattwa~Prasad Majumder, Katherine Hermann, Sean Welleck, Amir Yazdanbakhsh, and Peter Clark. 2023.
\newblock \href {https://openreview.net/forum?id=S37hOerQLB} {Self-refine: Iterative refinement with self-feedback}.
\newblock In \emph{Thirty-seventh Conference on Neural Information Processing Systems}.

\bibitem[{Mirchandani et~al.(2023)Mirchandani, Xia, Florence, brian ichter, Driess, Arenas, Rao, Sadigh, and Zeng}]{mirchandani2023large}
Suvir Mirchandani, Fei Xia, Pete Florence, brian ichter, Danny Driess, Montserrat~Gonzalez Arenas, Kanishka Rao, Dorsa Sadigh, and Andy Zeng. 2023.
\newblock \href {https://openreview.net/forum?id=RcZMI8MSyE} {Large language models as general pattern machines}.
\newblock In \emph{7th Annual Conference on Robot Learning}.

\bibitem[{Mirzadeh et~al.(2024)Mirzadeh, Alizadeh, Shahrokhi, Tuzel, Bengio, and Farajtabar}]{mirzadeh2024gsmsymbolicunderstandinglimitationsmathematical}
Iman Mirzadeh, Keivan Alizadeh, Hooman Shahrokhi, Oncel Tuzel, Samy Bengio, and Mehrdad Farajtabar. 2024.
\newblock \href {https://arxiv.org/abs/2410.05229} {Gsm-symbolic: Understanding the limitations of mathematical reasoning in large language models}.
\newblock \emph{Preprint}, arXiv:2410.05229.

\bibitem[{Odena et~al.(2021)Odena, Shi, Bieber, Singh, Sutton, and Dai}]{odena2021bustlebottomupprogramsynthesis}
Augustus Odena, Kensen Shi, David Bieber, Rishabh Singh, Charles Sutton, and Hanjun Dai. 2021.
\newblock \href {https://arxiv.org/abs/2007.14381} {Bustle: Bottom-up program synthesis through learning-guided exploration}.
\newblock \emph{Preprint}, arXiv:2007.14381.

\bibitem[{OpenAI(2024{\natexlab{a}})}]{gpt4omini}
OpenAI. 2024{\natexlab{a}}.
\newblock \href {https://openai.com/index/gpt-4o-mini-advancing-cost-efficient-intelligence/} {Gpt-4o mini: advancing cost-efficient intelligence}.

\bibitem[{OpenAI(2024{\natexlab{b}})}]{gpt4o}
OpenAI. 2024{\natexlab{b}}.
\newblock \href {https://openai.com/index/hello-gpt-4o/} {Hello gpt-4o}.

\bibitem[{Qiu et~al.(2024)Qiu, Jiang, Lu, Sclar, Pyatkin, Bhagavatula, Wang, Kim, Choi, Dziri, and Ren}]{qiu2024phenomenal}
Linlu Qiu, Liwei Jiang, Ximing Lu, Melanie Sclar, Valentina Pyatkin, Chandra Bhagavatula, Bailin Wang, Yoon Kim, Yejin Choi, Nouha Dziri, and Xiang Ren. 2024.
\newblock \href {https://openreview.net/forum?id=bNt7oajl2a} {Phenomenal yet puzzling: Testing inductive reasoning capabilities of language models with hypothesis refinement}.
\newblock In \emph{The Twelfth International Conference on Learning Representations}.

\bibitem[{Rule(2020)}]{rule2020child}
Joshua~S Rule. 2020.
\newblock \emph{The child as hacker: {{Building}} more human-like models of learning}.
\newblock Ph.D. thesis, MIT.

\bibitem[{Sablé-Meyer et~al.(2022)Sablé-Meyer, Ellis, Tenenbaum, and Dehaene}]{SABLEMEYER2022101527}
Mathias Sablé-Meyer, Kevin Ellis, Josh Tenenbaum, and Stanislas Dehaene. 2022.
\newblock \href {https://doi.org/10.1016/j.cogpsych.2022.101527} {A language of thought for the mental representation of geometric shapes}.
\newblock \emph{Cognitive Psychology}, 139:101527.

\bibitem[{Team(2025)}]{qwq}
Qwen Team. 2025.
\newblock \href {https://qwenlm.github.io/blog/qwq-32b/} {Qwq-32b: Embracing the power of reinforcement learning}.

\bibitem[{Tenenbaum et~al.(2011)Tenenbaum, Kemp, Griffiths, and Goodman}]{doi:10.1126/science.1192788}
Joshua~B. Tenenbaum, Charles Kemp, Thomas~L. Griffiths, and Noah~D. Goodman. 2011.
\newblock \href {https://doi.org/10.1126/science.1192788} {How to grow a mind: Statistics, structure, and abstraction}.
\newblock \emph{Science}, 331(6022):1279--1285.

\bibitem[{Tian et~al.(2020)Tian, Ellis, Kryven, and Tenenbaum}]{tian2020learning}
Lucas Tian, Kevin Ellis, Marta Kryven, and Josh Tenenbaum. 2020.
\newblock \href {https://proceedings.neurips.cc/paper_files/paper/2020/file/1c104b9c0accfca52ef21728eaf01453-Paper.pdf} {Learning abstract structure for drawing by efficient motor program induction}.
\newblock In \emph{Advances in Neural Information Processing Systems}, volume~33, pages 2686--2697. Curran Associates, Inc.

\bibitem[{Tu et~al.(2024)Tu, Li, Yu, Wang, Hou, and Li}]{tu2024chatlogcarefullyevaluatingevolution}
Shangqing Tu, Chunyang Li, Jifan Yu, Xiaozhi Wang, Lei Hou, and Juanzi Li. 2024.
\newblock \href {https://arxiv.org/abs/2304.14106} {Chatlog: Carefully evaluating the evolution of chatgpt across time}.
\newblock \emph{Preprint}, arXiv:2304.14106.

\bibitem[{Wang et~al.(2024)Wang, Zelikman, Poesia, Pu, Haber, and Goodman}]{wang2024hypothesis}
Ruocheng Wang, Eric Zelikman, Gabriel Poesia, Yewen Pu, Nick Haber, and Noah Goodman. 2024.
\newblock \href {https://openreview.net/forum?id=G7UtIGQmjm} {Hypothesis search: Inductive reasoning with language models}.
\newblock In \emph{The Twelfth International Conference on Learning Representations}.

\bibitem[{Wang et~al.(2023)Wang, Wei, Schuurmans, Le, Chi, Narang, Chowdhery, and Zhou}]{wang2023selfconsistency}
Xuezhi Wang, Jason Wei, Dale Schuurmans, Quoc~V Le, Ed~H. Chi, Sharan Narang, Aakanksha Chowdhery, and Denny Zhou. 2023.
\newblock \href {https://openreview.net/forum?id=1PL1NIMMrw} {Self-consistency improves chain of thought reasoning in language models}.
\newblock In \emph{The Eleventh International Conference on Learning Representations}.

\bibitem[{Wei et~al.(2022)Wei, Wang, Schuurmans, Bosma, brian ichter, Xia, Chi, Le, and Zhou}]{wei2022chain}
Jason Wei, Xuezhi Wang, Dale Schuurmans, Maarten Bosma, brian ichter, Fei Xia, Ed~H. Chi, Quoc~V Le, and Denny Zhou. 2022.
\newblock \href {https://openreview.net/forum?id=_VjQlMeSB_J} {Chain of thought prompting elicits reasoning in large language models}.
\newblock In \emph{Advances in Neural Information Processing Systems}.

\bibitem[{Wu et~al.(2024)Wu, Qiu, Ross, Aky{\"u}rek, Chen, Wang, Kim, Andreas, and Kim}]{wu-etal-2024-reasoning}
Zhaofeng Wu, Linlu Qiu, Alexis Ross, Ekin Aky{\"u}rek, Boyuan Chen, Bailin Wang, Najoung Kim, Jacob Andreas, and Yoon Kim. 2024.
\newblock \href {https://doi.org/10.18653/v1/2024.naacl-long.102} {Reasoning or reciting? exploring the capabilities and limitations of language models through counterfactual tasks}.
\newblock In \emph{Proceedings of the 2024 Conference of the North American Chapter of the Association for Computational Linguistics: Human Language Technologies (Volume 1: Long Papers)}, pages 1819--1862, Mexico City, Mexico. Association for Computational Linguistics.

\bibitem[{Yang et~al.(2024)Yang, Dong, Du, Cheng, Cambria, Liu, Gao, and Wei}]{yang-etal-2024-language}
Zonglin Yang, Li~Dong, Xinya Du, Hao Cheng, Erik Cambria, Xiaodong Liu, Jianfeng Gao, and Furu Wei. 2024.
\newblock \href {https://aclanthology.org/2024.eacl-long.13/} {Language models as inductive reasoners}.
\newblock In \emph{Proceedings of the 18th Conference of the European Chapter of the Association for Computational Linguistics (Volume 1: Long Papers)}, pages 209--225, St. Julian{'}s, Malta. Association for Computational Linguistics.

\bibitem[{Zhang et~al.(2023)Zhang, Li, Cui, Cai, Liu, Fu, Huang, Zhao, Zhang, Chen, Wang, Luu, Bi, Shi, and Shi}]{zhang2023sirenssongaiocean}
Yue Zhang, Yafu Li, Leyang Cui, Deng Cai, Lemao Liu, Tingchen Fu, Xinting Huang, Enbo Zhao, Yu~Zhang, Yulong Chen, Longyue Wang, Anh~Tuan Luu, Wei Bi, Freda Shi, and Shuming Shi. 2023.
\newblock \href {https://arxiv.org/abs/2309.01219} {Siren's song in the ai ocean: A survey on hallucination in large language models}.
\newblock \emph{Preprint}, arXiv:2309.01219.

\bibitem[{Zheng et~al.(2025{\natexlab{a}})Zheng, Chen, Li, Li, Zong, Shi, Xu, Song, Wong, and See}]{zheng2025cursecotlimitationschainofthought}
Tianshi Zheng, Yixiang Chen, Chengxi Li, Chunyang Li, Qing Zong, Haochen Shi, Baixuan Xu, Yangqiu Song, Ginny~Y. Wong, and Simon See. 2025{\natexlab{a}}.
\newblock \href {https://arxiv.org/abs/2504.05081} {The curse of cot: On the limitations of chain-of-thought in in-context learning}.
\newblock \emph{Preprint}, arXiv:2504.05081.

\bibitem[{Zheng et~al.(2025{\natexlab{b}})Zheng, Cheng, Li, Shi, Wang, Bai, Song, Wong, and See}]{zheng2025logidynamicsunravelingdynamicslogical}
Tianshi Zheng, Jiayang Cheng, Chunyang Li, Haochen Shi, Zihao Wang, Jiaxin Bai, Yangqiu Song, Ginny~Y. Wong, and Simon See. 2025{\natexlab{b}}.
\newblock \href {https://arxiv.org/abs/2502.11176} {Logidynamics: Unraveling the dynamics of logical inference in large language model reasoning}.
\newblock \emph{Preprint}, arXiv:2502.11176.

\bibitem[{Zhou et~al.(2024)Zhou, Tao, Zhu, Luo, Wang, and Han}]{zhou2024can}
Zhanke Zhou, Rong Tao, Jianing Zhu, Yiwen Luo, Zengmao Wang, and Bo~Han. 2024.
\newblock \href {https://openreview.net/forum?id=FbuODM02ra} {Can language models perform robust reasoning in chain-of-thought prompting with noisy rationales?}
\newblock In \emph{The Thirty-eighth Annual Conference on Neural Information Processing Systems}.

\end{thebibliography}
\newpage
\appendix

\section*{Appendices}
\section{Details on Evaluation Pipeline}
In this section, we provide detailed information on the evaluation pipeline, including the data construction and the performance assessment. 

For Arithmetic, we generate the base-7, base-8, and base-9 tasks by randomly sampling two two-digit numbers in the corresponding base, and then we check whether there is a carry-over in the addition process. If there is no carry-over, we regenerate the numbers. The noisy examples are generated in base-10. This choice reflects a common human-like calculation error\textemdash defaulting to the familiar decimal system\textemdash making it a ``meaningful'' and realistic form of noise in a mathematical context.

For Cryptography, we randomly select words of appropriate lengths from the NLTK Word Lists corpus\footnote{\url{https://www.nltk.org/nltk\_data/}}, and then we encrypt the words using the Caesar, Atbash, and Keyboard ciphers. The noisy examples are generated by randomly replacing the letters in the output with other letters. This mimics common errors in text processing or encryption, such as typographical mistakes or transmission noise, which are naturally occurring in symbolic domains.

For List Functions, we first write the corresponding rule functions for each task, and then we automatically generate the input data with appropriate lengths and ranges. The inputs are generated by randomly sampling numbers from a specific range with some constraints. Noise is introduced by randomly replacing some numbers in the output list with other values. This simulates natural errors, such as data corruption or misrecording, that might occur in sequence-based tasks.

During the data synthesis process, we attach manual supervision to ensure that the generated data can correctly induce the rules.
During the evaluation process, we use the inputs in the test set as the input for rule execution. We evaluate the model's performance using exact match. If the model's output is correct on all the test set examples, we consider the model to have successfully induced the rule. If the model fails to output a valid programmatic rule or the program contains an infinite loop or errors, we consider it a failure.

\label{sec:app_eval}
\section{Experimental Details}
\label{sec:app_exp}
\subsection{Experimental Settings}
For robustness under different noise levels, we run each experiment three times and report the mean and standard deviation of the results to avoid randomness. Except for the self-consistency (SC), hypothesis refinement (HR), and sample-steered rule refinement (SRR) that require diverse generations, we set the temperature to $0.0$ for all models to ensure reproducibility. For SC, HR and SRR, the temperature is set to $0.7$, consistent with the original work of SC~\cite{wang2023selfconsistency}. The positions of noise in seen examples are random to avoid positional bias~\cite{lu-etal-2022-fantastically}. In implementation, we choose $2$ subsets for SRR by splitting the seen examples into two subsets, and we generate $3$ hypotheses at each iteration of HR. The number of iterations is set to $3$ for SR, HR and SRR.

For the GPT-4o-mini and GPT-4o models, the experiments are conducted through the official OpenAI\footnote{\url{https://openai.com/api/}} API platform. For the DeepSeek-V3 and DeepSeek-R1 models, the experiments are conducted through the official DeepSeek\footnote{\url{https://platform.deepseek.com/usage}} API platform and the Volcengine\footnote{\url{https://www.volcengine.com/}} platform. We spent about $700$ USD in total for the experiments.

\begin{table}[htbp]
    \centering
    \small
    \begin{tabular}{c|ccc}
    \toprule
    \textbf{Method} & Acc$_{\text{clean}}$ & Acc$_{10\%~\text{noise}}$ & Consistency\\
    \midrule
    CoT & $66.0$ & $56.0$ & $64.5$ \\
    SC & $79.0$ & $72.0$ & $63.0$\\
    SR & $67.0$ & $60.0$ & $65.0$\\
    HR & $89.0$ & $76.0$ & $79.0$\\
    \midrule
    SRR & $100.0$ & $97.0$ & $97.0$\\
    \bottomrule
    \end{tabular}
    \caption{Task accuracy ($\%$) and consistency score ($\%$) of DeepSeek-V3 on the base-7 Arithmetic task with distribution noise.}
    \label{tab:noise_type}
\end{table}

\begin{table*}[ht]
    \centering
    \small
    \begin{adjustbox}{max width=1\linewidth}
    {
         \begin{tabular}{l|cccc|cccc|cccc|cccc|cccc|cccc|cccc}
         \toprule
         \multirow{2}{*}{\textbf{Method}} & \multicolumn{4}{c|}{Arithmetic$_7$} & \multicolumn{4}{c|}{Arithmetic$_8$} & \multicolumn{4}{c|}{Arithmetic$_9$} & \multicolumn{4}{c|}{Crypto$_{\text{Caesar}}$} & \multicolumn{4}{c|}{Crypto$_{\text{Atbash}}$} & \multicolumn{4}{c|}{Crypto$_{\text{Keyboard}}$} & \multicolumn{4}{c}{List Func.} \\
         \cmidrule{2-29}
         & \texttt{BR} & \texttt{BW} & \texttt{RW} & \texttt{WR} & \texttt{BR} & \texttt{BW} & \texttt{RW} & \texttt{WR} & \texttt{BR} & \texttt{BW} & \texttt{RW} & \texttt{WR}  & \texttt{BR} & \texttt{BW} & \texttt{RW} & \texttt{WR}  & \texttt{BR} & \texttt{BW} & \texttt{RW} & \texttt{WR}  & \texttt{BR} & \texttt{BW} & \texttt{RW} & \texttt{WR}  & \texttt{BR} & \texttt{BW} & \texttt{RW} & \texttt{WR}\\
         \midrule
         \rowcolor[gray]{0.9} \multicolumn{29}{c}{\textbf{GPT-4o-mini}} \\
         \midrule
         DO$_{10\%}$ & - & - & - & - & - & - & - & - & - & - & - & - & 64 & 182 & 22 & 32 & 15 & 258 & 4 & 23 & 0 & 289 & 0 & 11 & 190 & 492 & 45 & 23 \\
         DO$_{20\%}$ & - & - & - & - & - & - & - & - & - & - & - & - & 51 & 178 & 35 & 36 & 9 & 250 & 10 & 31 & 0 & 296 & 0 & 4 & 160 & 499 & 75 & 16 \\
         DO$_{30\%}$ & - & - & - & - & - & - & - & - & - & - & - & - & 38 & 188 & 48 & 26 & 14 & 247 & 5 & 34 & 0 & 290 & 0 & 10 & 126 & 493 & 109 & 22 \\
         \midrule
         \rowcolor[gray]{0.9} \multicolumn{29}{c}{\textbf{GPT-4o}} \\
         \midrule
         DO$_{10\%}$ & - & - & - & - & - & - & - & - & - & - & - & - & 195 & 64 & 10 & 31 & 85 & 137 & 59 & 19 & 7 & 282 & 7 & 4 & 259 & 433 & 40 & 18 \\
         DO$_{20\%}$ & - & - & - & - & - & - & - & - & - & - & - & - & 180 & 49 & 25 & 46 & 66 & 140 & 78 & 16 & 2 & 283 & 12 & 3 & 220 & 432 & 79 & 19 \\
         DO$_{30\%}$ & - & - & - & - & - & - & - & - & - & - & - & - & 155 & 56 & 50 & 39 & 49 & 137 & 95 & 19 & 3 & 282 & 11 & 4 & 178 & 433 & 121 & 18 \\
         \midrule
         CoT & 0 & 94 & 3 & 3 & 2 & 72 & 20 & 6 & 0 & 89 & 3 & 8 & 82 & 13 & 2 & 3 & 7 & 60 & 20 & 13 & 1 & 93 & 3 & 3 & 96 & 131 & 18 & 5 \\
         SC & 0 & 99 & 1 & 0 & 1 & 82 & 12 & 5 & 0 & 97 & 3 & 0 & 83 & 13 & 2 & 2 & 19 & 50 & 21 & 10 & 0 & 88 & 7 & 5 & 100 & 130 & 14 & 6 \\
         SR & 0 & 94 & 5 & 1 & 4 & 64 & 18 & 14 & 0 & 90 & 6 & 4 & 78 & 14 & 5 & 3 & 10 & 56 & 21 & 13 & 2 & 92 & 3 & 3 & 88 & 129 & 22 & 11 \\
         HR & 0 & 94 & 5 & 1 & 12 & 56 & 16 & 16 & 5 & 66 & 15 & 14 & 82 & 14 & 2 & 2 & 40 & 29 & 15 & 16 & 6 & 64 & 14 & 16 & 124 & 91 & 20 & 15 \\
         SRR & 0 & 89 & 6 & 5 & 23 & 27 & 22 & 28 & 4 & 64 & 17 & 15 & 81 & 14 & 1 & 4 & 39 & 32 & 16 & 13 & 0 & 83 & 8 & 9 & 136 & 96 & 11 & 7 \\
         \midrule
         \rowcolor[gray]{0.9} \multicolumn{29}{c}{\textbf{DeepSeek-V3}} \\
         \midrule
         DO$_{10\%}$ & - & - & - & - & - & - & - & - & - & - & - & - & 65 & 142 & 85 & 8 & 35 & 224 & 30 & 11 & 35 & 213 & 40 & 12 & 282 & 403 & 41 & 24 \\
         DO$_{20\%}$ & - & - & - & - & - & - & - & - & - & - & - & - & 37 & 141 & 113 & 9 & 37 & 208 & 28 & 27 & 24 & 215 & 51 & 10 & 238 & 400 & 85 & 27 \\
         DO$_{30\%}$ & - & - & - & - & - & - & - & - & - & - & - & - & 31 & 148 & 119 & 2 & 28 & 213 & 37 & 22 & 19 & 210 & 56 & 15 & 198 & 396 & 125 & 31 \\
         \midrule
         CoT & 141 & 19 & 26 & 14 & 160 & 2 & 32 & 6 & 110 & 12 & 53 & 25 & 156 & 26 & 13 & 5 & 29 & 114 & 34 & 23 & 5 & 168 & 22 & 5 & 236 & 184 & 56 & 24 \\
         SC & 76 & 7 & 10 & 7 & 92 & 0 & 7 & 1 & 69 & 4 & 15 & 12 & 84 & 11 & 3 & 2 & 25 & 42 & 18 & 15 & 1 & 83 & 10 & 6 & 129 & 91 & 19 & 11 \\
         SR & 58 & 8 & 22 & 12 & 66 & 9 & 17 & 8 & 45 & 15 & 17 & 23 & 64 & 10 & 18 & 8 & 5 & 66 & 15 & 14 & 2 & 83 & 9 & 6 & 105 & 94 & 38 & 13 \\
         HR & 74 & 6 & 12 & 8 & 82 & 2 & 9 & 7 & 67 & 11 & 11 & 11 & 84 & 12 & 2 & 2 & 40 & 28 & 16 & 16 & 4 & 63 & 20 & 13 & 153 & 70 & 21 & 6 \\
         SRR & 93 & 0 & 4 & 3 & 94 & 0 & 5 & 1 & 88 & 0 & 6 & 6 & 84 & 11 & 3 & 2 & 37 & 32 & 16 & 15 & 2 & 78 & 11 & 9 & 154 & 73 & 15 & 8 \\
         \bottomrule
    \end{tabular}
    }
    \end{adjustbox}
   
    \caption{Detailed breakdown of the consistency score of different methods or different noise levels. The notation \texttt{BR}, \texttt{BW}, \texttt{RW}, and \texttt{WR} represent both right, both wrong, right to wrong from clean condition to noisy condition, and wrong to right from clean condition to noisy condition, respectively.}
    \label{tab:breakdown_consis}
\end{table*}

\subsection{Additional Results}
\paragraph{Different Types of Noise} To ensure the noise reflects perturbations encountered in real-world scenarios, the noise of Arithmetic is different from that of Cryptography and List Functions. To explore the consistency of the results under a more uniform noise strategy, we conducted additional experiments on the base-7 Arithmetic task using DeepSeek-V3. Instead of base-10 outputs, we introduce disturbance noise by adding a random number between $1$ and $6$ (base\_num$-1$) to the correct results, aligning this noise more closely with the random corruption used in other tasks. The results are presented in Table~\ref{tab:noise_type}.

From the results, we find that the performance of CoT, SC, SR and SRR exhibits the same trend as in the original experiments, with a consistent drop in accuracy when noise is introduced. The consistency score also shows a similar trend, indicating the robustness fragility of inductive reasoning in the presence of noise. The consistency drop is more evident than the accuracy drop, suggesting the accuracy drop is not solely due to the failure of previous solvable cases. This further supports the need to consider a more detailed granularity except for the accuracy. 

\paragraph{Breakdown of Consistency Score} For a more comprehensive understanding of the consistency score, we provide a detailed breakdown of the consistency score of the experiments in Table~\ref{tab:breakdown_consis}. From the results, we observe that the introduction of noise not only leads to a performance drop in the original solvable cases but also results in some previously unsolved cases being solved. This reveals the fragility and inconsistency in the rule induction process. 

\paragraph{Ablation} We conduct ablation studies to investigate the impact of sampling right and wrong examples during the refinement process of SRR. The results are shown in Table~\ref{tab:ablation_sample}. From the results, we observe that sampling right and wrong examples during the refinement process of SRR has a positive impact on the performance when noise is introduced, while it has a slightly negative impact on the performance in the clean condition. This suggests that sampling can help the model to better adapt to the noise, the performance gap between the clean and noisy conditions is reduced.

\begin{table}[htbp]
    \centering
    \small
    \begin{adjustbox}{max width=1.0\linewidth}
    {
    \begin{tabular}{l|cc|cc}
    \toprule
    \multirow{2}{*}{\textbf{Dataset}} & \multicolumn{2}{c|}{{\textbf{Accuracy}$_{\text{clean}}$}} & \multicolumn{2}{c}{\textbf{Accuracy}$_{10\%~\text{noise}}$} \\
    \cmidrule{2-5}
    & \texttt{w/sample} & \texttt{w/o sample} & \texttt{w/sample} & \texttt{w/o sample} \\
    \midrule
    Arithmetic$_7$ & $97.0$ & $99.0$ & $96.0$ & $91.0$ \\
    Arithmetic$_8$ & $99.0$ & $98.0$ & $95.0$ & $96.0$ \\
    Arithmetic$_9$ & $94.0$ & $94.0$ & $94.0$ & $88.0$ \\
    Crypto$_{\text{Cae}}$ & $87.0$ & $86.0$ & $86.0$ & $85.0$ \\
    Crypto$_{\text{Atb}}$ & $53.0$ & $51.0$ & $52.0$ & $44.0$ \\
    Crypto$_{\text{Key}}$ & $13.0$ & $18.0$ & $11.0$ & $19.0$ \\
    List Func. & $67.6$ & $66.8$ & $64.8$ & $64.4$ \\
    \midrule
    \textbf{Average} & $72.9$ & $73.3$ & $71.3$ & $69.6$ \\
    \bottomrule
    \end{tabular}
    }
    \end{adjustbox}
    \caption{The impact of sampling right and wrong examples during the refinement process of SRR using DeepSeek-V3. The results are reported in terms of task accuracy ($\%$). \texttt{w/sample} and \texttt{w/o sample} represent with and without sampling, respectively.}
    \label{tab:ablation_sample}
\end{table}

\begin{figure*}[htbp]
    \centering
    \includegraphics[width=1.0\linewidth]{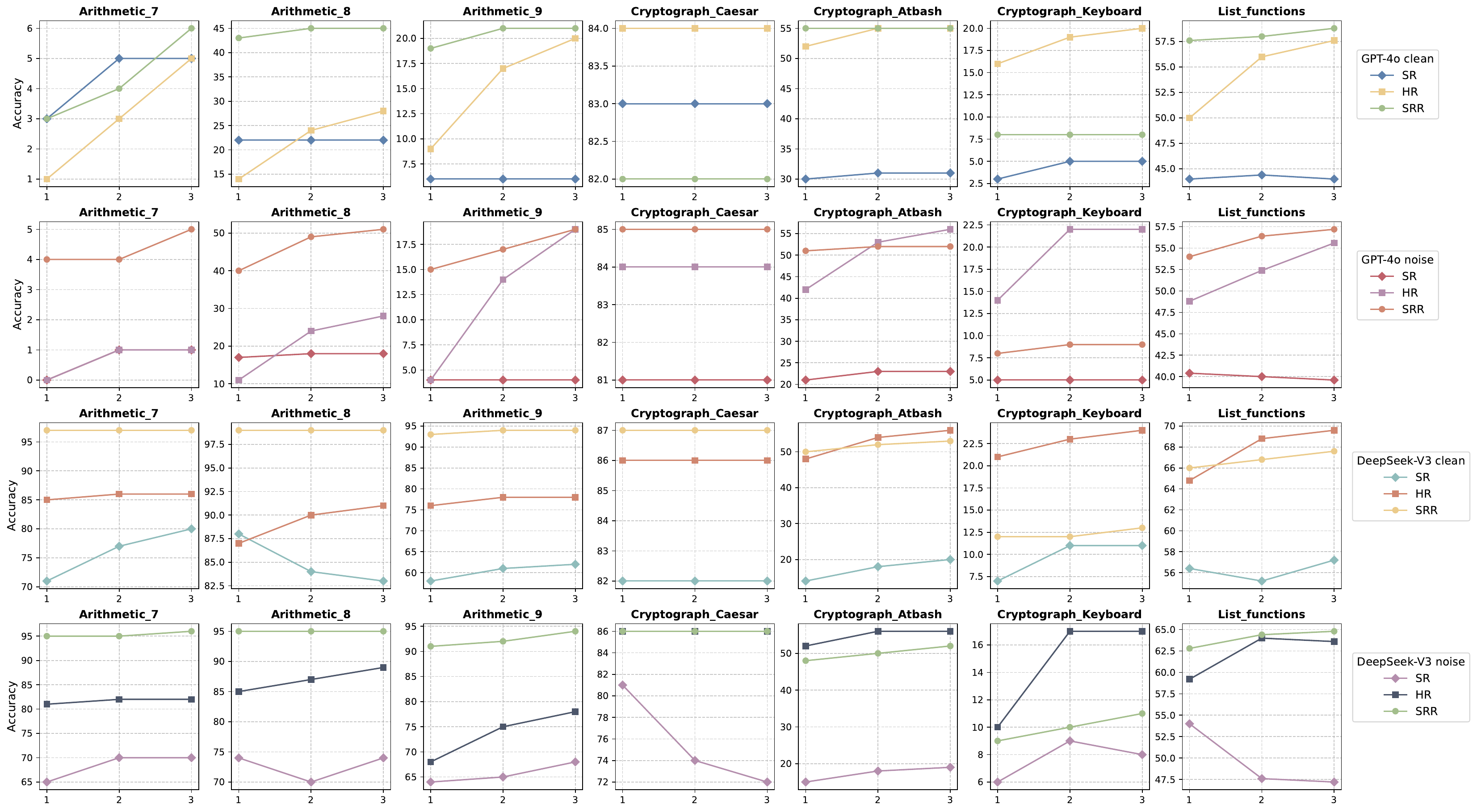}
    \caption{The performance of SR, HR and SRR across different iterations with different models and noise levels. The results are reported in terms of task accuracy ($\%$).}
    \label{fig:iteration}
\end{figure*}
% Ablations: with sample, without sample, performane across iteration
We also investigate the performance of SR, HR, and SRR across different iterations. The results are shown in Figure~\ref{fig:iteration}. Compared to the performance of SRR and HR, the performance of SR is not stable across different refinement iterations, and even shows a performance drop in some cases. This indicates that the LLM-generated feedback in SR is not always helpful for the model to refine the rule induction process. In contrast, both HR and SRR show a more stable performance improvement across iterations, and the improvement is more pronounced in HR. This is because HR generates different refined hypotheses at the same iteration and selects the best one, while SRR only refines the existing hypothesis. Specifically, HR also shows a performance drop in the DeepSeek-V3 model on the List Functions task with $10\%$ noise, which is likely due to HR only selecting the best hypothesis from the generated ones at each iteration. At the same time, SRR chooses the best one from all the generated rules.

% computational cost
\paragraph{Computational Cost} The computational cost of different methods is an important factor to consider the effectiveness, which is measured by the number of output tokens in the whole process. In general, methods that require more samples or generate more responses tend to have higher computational costs. We summarize the computational cost of different methods in Table~\ref{tab:token_cost}.

\begin{table}[htbp]
    \centering
    \small
    \begin{adjustbox}{max width=1.0\linewidth}
    {
    \begin{tabular}{l|cc|cc}
    \toprule
    \textbf{Method} & DeepSeek-V3$_\text{clean}$ & DeepSeek-V3$_\text{noise}$ & GPT-4o$_\text{clean}$ & GPT-4o$_\text{noise}$ \\
    \midrule
    CoT & $1,120$ & $1,194$ & $828$ & $800$ \\
    SC & $5,271$ & $5,620$ & $3,719$ & $3,709$ \\
    SR & $2,333$ & $2,673$ & $1,289$ & $1,192$ \\
    HR & $4,261$ & $4,884$ & $4,089$ & $4,211$ \\
    SRR & $3,388$ & $3,574$ & $2,791$ & $2,879$ \\
    \bottomrule
    \end{tabular}
    }
    \end{adjustbox}
    \caption{The computational cost of different methods. The results are reported in terms of the average number of output tokens per instance per dataset.}
    \label{tab:token_cost}
\end{table}
% reasoning models
\paragraph{More Reasoning Models} Except for the reasoning model used in~\ref{subsec:ees}, we also conduct experiments with QwQ-32B, another reasoning model implementing a reinforcement learning scaling approach~\cite{qwq}. Our goal is to explore whether the introduced noise can impact the performance of the reasoning models. The results based on Chain-of-Thought are shown in Table~\ref{tab:more_reasoning_models}.
Even the performance of reasoning models is more powerful than the LLMs, they still suffer from the noise and counterfactual tasks. 
\begin{table}[htbp]
    \centering
    \small
    \begin{adjustbox}{max width=1.0\linewidth}
    {
    \begin{tabular}{l|ccc|ccc}
    \toprule
    \multirow{2}{*}{\textbf{Dataset}} & \multicolumn{3}{c|}{\textbf{DeepSeek-R1}} & \multicolumn{3}{c}{\textbf{QwQ-32B}} \\
    \cmidrule{2-7}
    & Acc$_{\text{clean}}$ & Acc$_{\text{noise}}$ & Consistency & Acc$_{\text{clean}}$ & Acc$_{\text{noise}}$ & Consistency \\
    \midrule
    Arithmetic$_7$ & $99.0$ & $96.0$ & $95.0$ & $100.0$ & $99.0$ & $99.0$ \\
    Arithmetic$_8$ & $100.0$ & $99.0$ & $99.0$ & $100.0$ & $100.0$ & $100.0$ \\
    Arithmetic$_9$ & $98.0$ & $98.0$ & $96.0$ & $100.0$ & $99.0$ & $99.0$ \\
    \midrule
    Crypto$_\text{Caesar}$ & $84.0$ & $83.0$ & $93.0$ & $77.0$ & $75.0$ & $84.0$ \\
    Crypto$_\text{Atbash}$ & $65.0$ & $65.0$ & $70.0$ & $46.0$ & $41.0$ & $69.0$ \\
    Crypto$_\text{Key}$ & $2.0$ & $0.0$ & $98.0$ & $0.0$ & $0.0$ & $100.0$ \\
    \midrule
   List Func. & $76.0$ & $68.4$ & $87.6$ & $72.8$ & $68.0$ & $85.6$ \\
    \bottomrule
    \end{tabular}
    }
    \end{adjustbox}
    \caption{Task accuracy ($\%$) and consistency score of DeepSeek-R1 and QwQ-32B with Chain-of-Thought prompting on different datasets.}
    \label{tab:more_reasoning_models}
\end{table}

\section{Prompt Templates}
We provide all the prompts we used in this section. For the self-refine, hypothesis refinement, and sample-steered rule refinement setting, we use the chain-of-thought prompt as the initial prompt; the iterative prompts are different due to the type of feedback. 

\onecolumn
\begin{tcolorbox}[title={\textsc{\textbf{Prompt for Direct Output}}}, colback=pback, colframe=pframe, coltitle=white,center title]
\tt
Please generate a rule that maps the following inputs to their corresponding outputs using a Python function. The input is \textbf{\{Input Description\}}. The output is \textbf{\{Output Description\}}. Note that some examples may be wrong, and you should take this into account when proposing the rule. \\
\\
    \textbf{\{examples\}}\\
\\
    Please format your Python function as follows:\\
    \verb|```|python\\
    def fn(x):\\
    \quad\# Your code here\\
    \verb|```|\\
    Your response should only include the function definition, not the function call or any other information.
\end{tcolorbox}

\begin{tcolorbox}[title={\textsc{\textbf{Prompt for Chain-of-Thought}}}, colback=pback, colframe=pframe, coltitle=white,center title]
\tt
Please generate a rule that maps the following inputs to their corresponding outputs using a Python function. The input is \textbf{\{Input Description\}}. The output is \textbf{\{Output Description\}}. Note that some examples may be wrong, and you should take this into account when proposing the rule. \\
\\
    \textbf{\{examples\}}\\
\\
    Please format your Python function as follows:\\
    \verb|```|python\\
    def fn(x):\\
    \quad\# Your code here\\
    \verb|```|\\
    Think step-by-step and explain your reasoning. Your response should include your thought process and the function definition without the function call.
\end{tcolorbox}

\begin{tcolorbox}[title={\textsc{\textbf{Prompt for Self Refine}}}, colback=pback, colframe=pframe, coltitle=white,center title]
\tt
\begin{center}
    \textbf{\underline{Feedback Generation}}
\end{center}
You have generated a rule that maps the following inputs to their corresponding outputs using a Python function. The input is \textbf{\{Input Description\}}. The output is \textbf{\{Output Description\}}.\\
\\
    \textbf{\{examples\}}\\
\\
    In the last step, your rule is:\\
\\
    \verb|```|python\\
    \textbf{\{rule\}}\\
    \verb|```|\\
\\
    Give some feedback on the rule you have generated, like how it can be improved, what is wrong with it, etc.\\
    Your response should only include the feedback. If you think the rule is good enough, your response should be ``NO FEEDBACK'' without other information. Note that some examples may be wrong, and you should take this into account when proposing the feedback.\\
\begin{center}
    \textbf{\underline{Iteration}}
\end{center}
You have generated a rule that maps the following inputs to their corresponding outputs using a Python function. The input is \textbf{\{Input Description\}}. The output is \textbf{\{Output Description\}}. Note that some examples may be wrong, and you should take this into account when proposing the rule.\\
\\
    \textbf{\{examples\}} \\
\\
    In the last step, your rule is:\\
\\
    \verb|```|python\\
    \textbf{\{rule\}}\\
    \verb|```|\\
\\
    The feedback you have given is:\\
\\
    \textbf{\{feedback\}}\\
\\
    Generate a new rule that maps the given inputs to their corresponding outputs using a Python function. Please format your rule as follows:\\
    \verb|```|python\\
    def fn(x):\\
    \quad\# Your code here\\
    \verb|```|\\
    Think step-by-step and explain your reasoning. Your response should include your thought process and the function definition without the function call.
\end{tcolorbox}

\begin{tcolorbox}[title={\textsc{\textbf{Prompt for Iteration in Hypothesis Refinement}}}, colback=pback, colframe=pframe, coltitle=white,center title]
\tt
You have generated a rule that maps the following inputs to their corresponding outputs using a Python function. The input is \textbf{\{Input Description\}}. The output is \textbf{\{Output Description\}}. Note that some examples may be noisy, and you should take this into account when proposing the rule.\\
\\
\textbf{\{examples\}}\\
\\
In the last step, your rule is\\
\\
\verb|```|python\\
\textbf{\{rule\}}\\
\verb|```|\\
\\
It does not work for the following examples:\\
\\
\textbf{\{wrong examples\}}\\
\\
Generate a new rule that maps the given inputs to their corresponding outputs using a Python function. Please format your rule as follows:\\
\verb|```|python\\
def fn(x):\\
\quad\# Your code here\\
\verb|```|\\
Think step-by-step and explain your reasoning. Your response should include your thought process and the function definition without the function call. You can either modify the existing rule or propose a new one.
\end{tcolorbox}

\begin{tcolorbox}[title={\textsc{\textbf{Prompt for Iteration in Sample-steered rule refinement}}}, colback=pback, colframe=pframe, coltitle=white,center title]
\tt
You have generated a rule that maps the following inputs to their corresponding outputs using a Python function. The input is \textbf{\{Input Description\}}. The output is \textbf{\{Output Description\}}. Note that some examples may be noisy, and you should take this into account when proposing the rule. In the last step, your rule is\\
\\
    \verb|```|python\\
    \textbf{\{rule\}}\\
    \verb|```|\\
\\
    But this rule is not correct. It works for the following examples:\\
\\
    \textbf{\{sampled right examples\}}\\
\\
    However, it does not work for the following examples:\\ 
\\
    \textbf{\{sampled wrong examples\}}\\
\\
    Generate a new rule that maps the given inputs to their corresponding outputs using a Python function. Please format your rule as follows:\\
    \verb|```|python\\
    def fn(x):\\
    \quad\# Your code here\\
    \verb|```|\\
    Think step-by-step and explain your reasoning. Your response should include your thought process and the function definition without the function call. You can either modify the existing rule or propose a new one.
\end{tcolorbox}

\end{document}